\documentclass{article}

\PassOptionsToPackage{numbers,sort&compress}{natbib}
\usepackage[main,preprint]{neurips_2026}

\usepackage{placeins}
\usepackage{needspace}
\usepackage[utf8]{inputenc} 
\usepackage[T1]{fontenc}    
\usepackage{hyperref}       
\usepackage{url}            
\usepackage{booktabs}       
\usepackage{amsfonts}       
\usepackage{nicefrac}       
\usepackage{microtype}      
\usepackage[table]{xcolor}  
\usepackage{graphicx}
\usepackage{graphicx}
\usepackage{subcaption}
\usepackage{float}
\usepackage{graphicx}
\usepackage{subcaption} 
\usepackage{multirow}
\usepackage{caption}
\makeatletter
  
\usepackage{bm}   
\usepackage{graphicx}

\usepackage{array}
\usepackage{tabularx}
\usepackage[table]{xcolor}
\usepackage{booktabs}
\usepackage{multirow}
\usepackage{makecell}
\usepackage{array}
\usepackage{caption}
\usepackage{subcaption}

\usepackage{wrapfig}
\newcolumntype{Y}{>{\centering\arraybackslash}X}
\usepackage{amsmath}
\usepackage{wrapfig}
\usepackage[table]{xcolor}
\usepackage{makecell}
\definecolor{sectiongray}{RGB}{232,232,232}
\definecolor{oursbg}{RGB}{247,243,221}
\definecolor{bestred}{RGB}{190,30,45}

\newcommand{\best}[1]{\textcolor{bestred}{\textbf{#1}}}

\usepackage{amssymb}
\usepackage[table]{xcolor}
\usepackage{makecell}
\usepackage{booktabs} 
\usepackage{caption}
\usepackage{float}
\usepackage{stfloats}
\usepackage{caption}
\usepackage{float}
\usepackage{hyperref}
\usepackage{algorithm}
\usepackage{algpseudocode}
\usepackage{cuted}
\usepackage{caption}

\algrenewcommand\algorithmicrequire{\textbf{Input:}}
\algrenewcommand\algorithmicensure{\textbf{Output:}}
\algrenewcommand\algorithmicreturn{\textbf{return}}

\usepackage{physics}
\usepackage{mathtools}

\title{When and What to Prune? 
Stage-Aware Visual Token Pruning for Efficient VLA}

\author{%
  Tianjun Shi\textsuperscript{1}, Haotian Xiong\textsuperscript{1}, Ziyu Gong\textsuperscript{1}, Qi Lu\textsuperscript{1}, Li Li\textsuperscript{1}\thanks{Corresponding author.}\\
  \textsuperscript{1}University of Macau\\
  \texttt{\{mc55413,mc55441,mc55143,yc57960,llili\}@um.edu.mo}
}

\begin{document}

\raggedbottom

\maketitle
\vspace{-18pt}

\begin{abstract}
Visual token pruning is an effective way to accelerate vision-language models and is especially useful for vision-language-action (VLA) inference, where many visual tokens must be processed before predicting robot actions. Existing pruning methods usually estimate which tokens can be pruned based on attention scores or feature diversity, retaining tokens that are either highly attended or visually different from others. However, most of them use fixed pruning schedules, such as pruning once at a preset layer or pruning at uniformly spaced layers. Such schedules can be risky for VLA models, because the model may not know which visual regions matter for the action in early layers. Tokens that look unimportant at first may become useful after the model combines visual observations with the language instruction. In this work, we propose \textbf{SAPrune}, a training-free visual token pruning framework for efficient VLA inference. Instead of pruning at fixed layers, SAPrune uses a small calibration set to observe how action-to-visual attention changes across layers, and chooses pruning layers only after the attention pattern becomes more reliable. At each selected layer, SAPrune applies a dual-path pruning rule: one path protects strongly attended visual tokens from pruning, while the other prevents useful surrounding context from being discarded. Experiments on LIBERO, SIMPLER, and real-world robotic tasks show that SAPrune prunes 87.5\% of visual tokens and achieves up to \textbf{1.718$\times$} inference speedup while maintaining competitive task success rates.
\end{abstract}

\section{Introduction}

Vision-language-action (VLA) models have recently shown strong potential for embodied decision making by mapping visual observations and language instructions directly to robot actions~\citep{brohan2023rt1,zitkovich2023rt2,oneill2024openx,octo2024,kim2025openvla,black2024pi0}. Despite their effectiveness, VLA inference remains computationally expensive because a large number of visual tokens must be propagated through deep Transformer layers~\citep{kim2025openvla,xu2025vlacache,yang2025efficientvla,liu2025vlapruner}. Visual token pruning is therefore a natural direction for reducing latency and computation, as shown in prior studies on efficient Vision Transformers and large vision-language models~\citep{rao2021dynamicvit,liang2022evit,bolya2023tome,chen2024fastv,huang2024ivtp,zhang2025sparsevlm,sun2025lvpruning,alvar2025divprune}. 

\begin{figure}[t]
    \centering
    \includegraphics[width=0.95\linewidth]{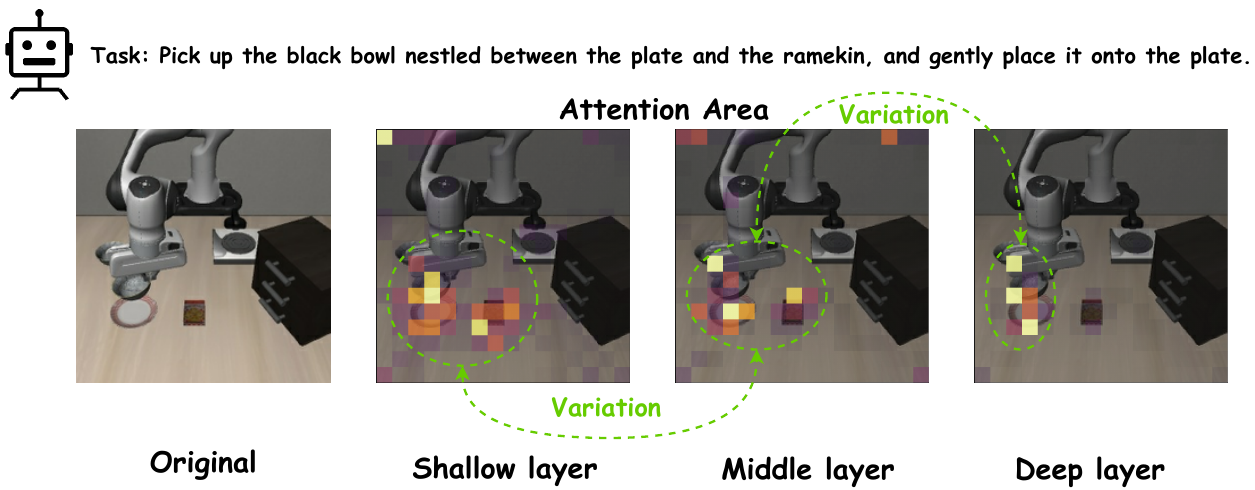}
    \caption{Layer-wise visualization of attention in VLA.}
    \label{fig:attention_dynamics}
\end{figure}

Although recent studies have shown substantial redundancy in VLA inference, most pruning-oriented acceleration methods still decide token removal from local importance cues, visual saliency, feature-space diversity, or their fixed combinations~\citep{chen2024fastv,zhang2025sparsevlm,huang2024ivtp,sun2025lvpruning,zhang2025vispruner,alvar2025divprune,yang2025efficientvla,liu2025vlapruner}. This treats pruning mainly as a static token-selection problem. Such a formulation is insufficient for action generation, where visual evidence does not remain fixed across depth: it gradually emerges, redistributes, and stabilizes as the model forms an action decision. Moreover, visual distinctiveness alone does not imply decision usefulness, since tokens that are geometrically diverse may still be irrelevant to the current action. These limitations motivate us to reconsider VLA pruning from two questions: \emph{when} pruning becomes reliable, and \emph{what} action-relevant evidence should be preserved.

\begin{figure}[!htbp]
    \centering
    \begin{subfigure}[t]{0.56\linewidth}
        \centering
        \includegraphics[width=\linewidth]{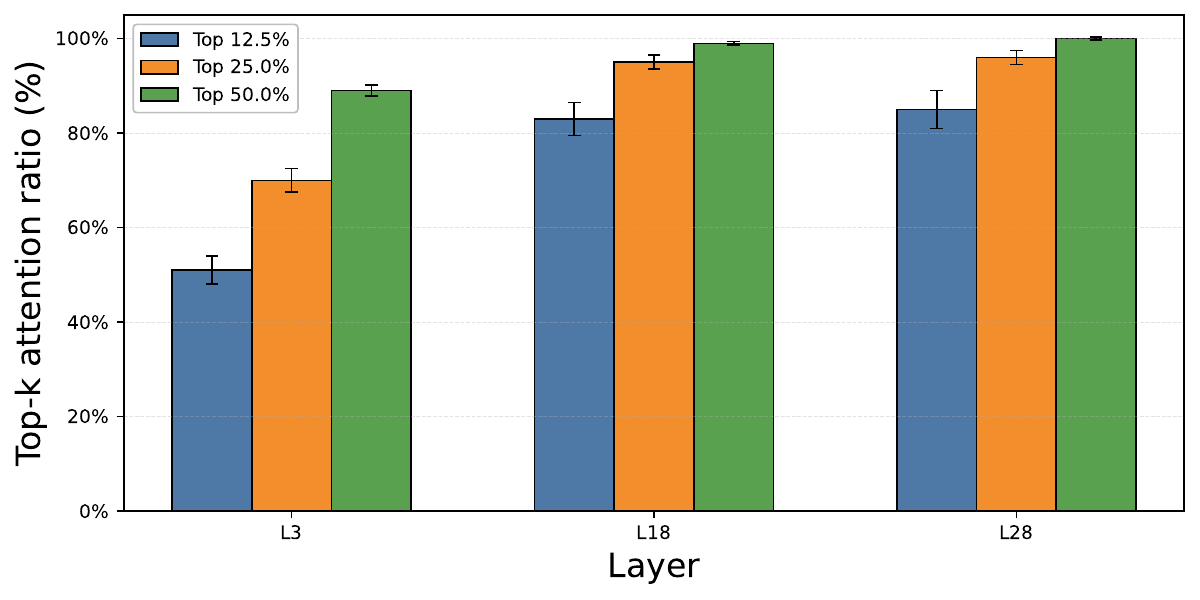}
        \caption{Visualization of Top-k Importance Image Token Received Attention Ratio.}
        \label{fig:obs2}
    \end{subfigure}
    \hfill
    \begin{subfigure}[t]{0.43\linewidth}
        \centering
        \includegraphics[width=\linewidth]{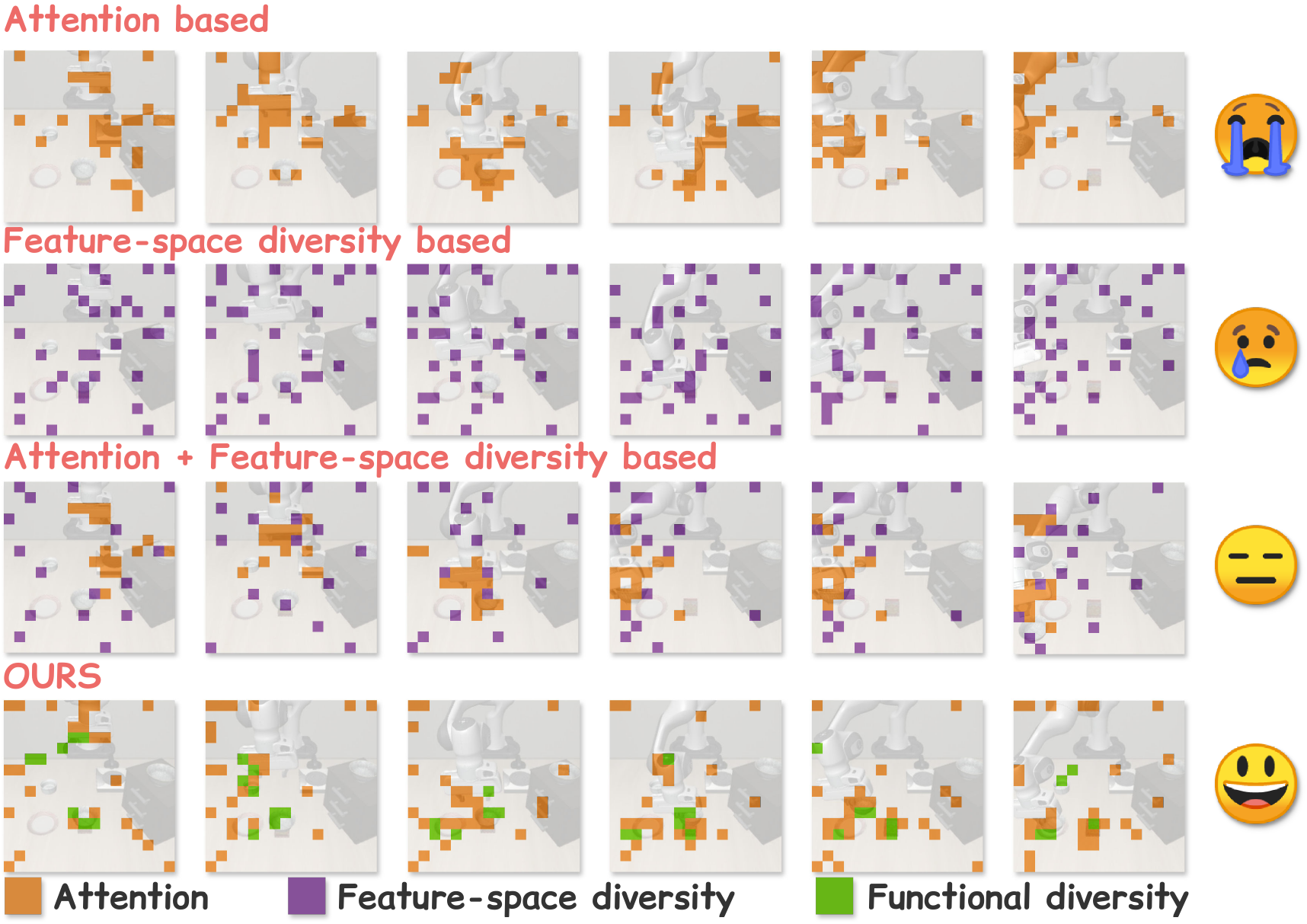}
        \caption{Comparison of Performance under Different Token Pruning Strategies.}
        \label{fig:obs3}
    \end{subfigure}
    \caption{Preliminary observations on top-k attention concentration and token pruning strategy comparison.}
    \label{fig:observations}
\end{figure}

To understand when visual tokens can be safely pruned during VLA inference, we analyze the attention from action-related states to visual tokens across Transformer layers. Figure~\ref{fig:attention_dynamics} shows that early layers attend to a broad set of visual tokens, while middle and deep layers gradually concentrate on tokens related to the target action. Figure~\ref{fig:obs2} further quantifies this trend using the top-$k$ attention mass: the mass increases quickly from shallow to middle layers and then changes more slowly. This suggests that early layers have not yet clearly determined which visual tokens are useful for the action. Therefore, pruning too early may remove tokens that become important later, motivating us to delay pruning until the attention pattern becomes more stable. We further compare different pruning criteria in Fig.~\ref{fig:obs3}. Attention-only selection tends to concentrate on dominant regions and may miss supportive context, whereas feature-space diversity-based selection may retain tokens that are visually distinctive but weakly related to the current action. These observations suggest that attention-only selection, feature-space diversity-based selection, and their fixed combination are all insufficient for aggressive VLA pruning. 
Based on these principles, we propose \textbf{SAPrune}, a training-free visual token pruning framework for efficient VLA inference. SAPrune first selects pruning layers from rollout-level action-attention dynamics, and then applies a dual-path retention mechanism at each selected layer. The attention-core path preserves directly action-relevant tokens, while the functional-diversity path selects complementary tokens using query-conditioned functional signatures. Experiments on LIBERO~\citep{liu2023libero}, SIMPLER~\citep{li2024simpler}, and real-world robotic tasks show that SAPrune retains only 32 visual tokens, prunes 87.5\% of the original visual tokens, and achieves up to 1.718$\times$ speedup while maintaining competitive task success rates.

\section{Preliminary}

\subsection{Related Works}

\noindent\textbf{Efficient inference for VLA models.}
Vision-Language-Action (VLA) models extend large vision-language models by introducing an action modality, enabling robots to predict executable actions from visual observations and language instructions. Representative generalist robot policies include RT-1~\citep{brohan2023rt1}, RT-2~\citep{zitkovich2023rt2}, Open X-Embodiment / RT-X~\citep{oneill2024openx}, Octo~\citep{octo2024}, OpenVLA~\citep{kim2025openvla}, $\pi_0$~\citep{black2024pi0}. Although effective for general-purpose manipulation, these models repeatedly process dense visual tokens through deep Transformer layers, making inference costly for real-time robot deployment~\citep{xu2025vlacache,yang2025efficientvla,li2025spvla,pei2025vlaadp,liu2025vlapruner}.

\noindent\textbf{Visual token pruning for VLMs.}
Visual token pruning reduces visual redundancy by retaining a compact subset of informative visual tokens. Early efficient Vision Transformer methods, such as DynamicViT~\citep{rao2021dynamicvit}, EViT~\citep{liang2022evit}, and ToMe~\citep{bolya2023tome}, reduce computation through token sparsification or merging. For large vision-language models, attention-based methods such as FastV~\citep{chen2024fastv}, SparseVLM~\citep{zhang2025sparsevlm}, FoPru~\citep{jiang2024fopru}, HiPrune~\citep{liu2025hiprune}, LVPruning~\citep{sun2025lvpruning}, IVTP~\citep{huang2024ivtp}, and ZSPA-Prune~\citep{zhang2025zspaprune} rank tokens by prefill attention, text-to-vision attention, or prompt-conditioned relevance. Other methods preserve diversity or global representativeness~\citep{alvar2025divprune,zhang2025vispruner,zou2025holov,chen2026otprune}, while calibration or layer-adaptive methods determine pruning layers, ratios, or progressive schedules from calibration statistics or cross-layer information flow~\citep{ye2025fitprune,lin2024vtw,lee2025tamp,li2025btp,yang2025vflowopt}. However, these methods are mainly designed for VLM understanding and do not explicitly model action-dependent visual evidence in VLA inference.

\noindent\textbf{Training-free acceleration for VLA models.}
Training-free VLA acceleration mainly follows temporal reuse, adaptive computation, and pruning-based acceleration. VLA-Cache~\citep{xu2025vlacache} reuses visual features across adjacent timesteps, while VLA-ADP~\citep{pei2025vlaadp} dynamically reduces computation during manipulation. EfficientVLA~\citep{yang2025efficientvla}, SP-VLA~\citep{li2025spvla}, SpecPrune-VLA~\citep{wang2025specprunevla}, and VLA-Pruner~\citep{liu2025vlapruner} reduce visual-token computation during VLA inference. These methods reveal substantial redundancy in VLA models, but often rely on fixed pruning locations, semantic relevance, temporal reuse, or additional acceleration modules. They leave underexplored how to choose pruning layers and preserve action-relevant evidence under aggressive token reduction.
\subsection{VLA Inference and Action-to-Visual Attention}
\label{sec:vla_prelim}

\noindent\textbf{VLA inference.}
Vision-language-action (VLA) models map visual observations and language instructions to executable robot actions. At rollout step $t$, the model takes an image observation $I_t$ and a language instruction $x_t$ as input, and predicts an action $a_t$. For notational simplicity, we omit the timestep index in the following. The vision encoder produces $N$ visual tokens $V^{(0)}=\{v^{(0)}_1,\ldots,v^{(0)}_N\}$, while the text encoder produces $M$ text tokens $X^{(0)}=\{x^{(0)}_1,\ldots,x^{(0)}_M\}$. Depending on the backbone, action-related states can be implemented as action tokens, action placeholders, or action query states, denoted by $Q^{(0)}_a$, as in OpenVLA~\citep{kim2025openvla}, $\pi_0$~\citep{black2024pi0}. The initial multimodal sequence is
\begin{equation}
H^{(0)} = V^{(0)} \oplus X^{(0)} \oplus Q^{(0)}_a ,
\end{equation}
where $\oplus$ denotes sequence concatenation. The sequence is then propagated through $L$ Transformer layers:
\begin{equation}
H^{(l)} = F^{(l)}(H^{(l-1)}), 
\qquad
H^{(l)} = V^{(l)} \oplus X^{(l)} \oplus Q^{(l)}_a .
\end{equation}
The final robot action is predicted from the action-related output states. Since visual tokens usually occupy a large portion of the multimodal sequence, reducing $V^{(l)}$ is an effective way to lower VLA inference cost~\citep{xu2025vlacache,yang2025efficientvla,li2025spvla,pei2025vlaadp,liu2025vlapruner}.

\noindent\textbf{Action-to-visual attention.}
For visual token pruning, we use the attention from action-related states to visual tokens as the decision-conditioned relevance signal. At layer $l$, the action-to-visual attention score of visual token $i$ is defined as
\begin{equation}
a^{(l)}_i =
\frac{1}{|Q^{(l)}_a|H}
\sum_{q\in Q^{(l)}_a}\sum_{h=1}^{H}
\alpha^{(l,h)}_{q,v^{(l)}_i},
\label{eq:action_visual_attention}
\end{equation}
where $\alpha^{(l,h)}_{q,v_i}$ denotes the attention weight from action-related query $q$ to visual token $v_i$ in head $h$, $H$ is the number of attention heads, and $Q^{(l)}_a$ is the set of action-related states.

To characterize the concentration of action-relevant visual evidence at layer $l$, we compute the top-$\kappa$ attention mass:
\begin{equation}
S_l(\kappa)=
\sum_{i\in \mathrm{Top}_{\lfloor \kappa N\rfloor}(a^{(l)})}
a^{(l)}_i ,
\label{eq:topk_attention_mass}
\end{equation}
where $\mathrm{Top}_{\lfloor \kappa N\rfloor}(a^{(l)})$ denotes the indices of the top $\lfloor \kappa N\rfloor$ visual tokens ranked by action-to-visual attention. A larger $S_l(\kappa)$ indicates that action evidence is concentrated on fewer visual tokens. We use this signal for both pruning-layer selection and action-relevant token retention.

\section{Method}

\subsection{Overview}

\begin{figure}[t]
    \centering
    \includegraphics[width=\linewidth]{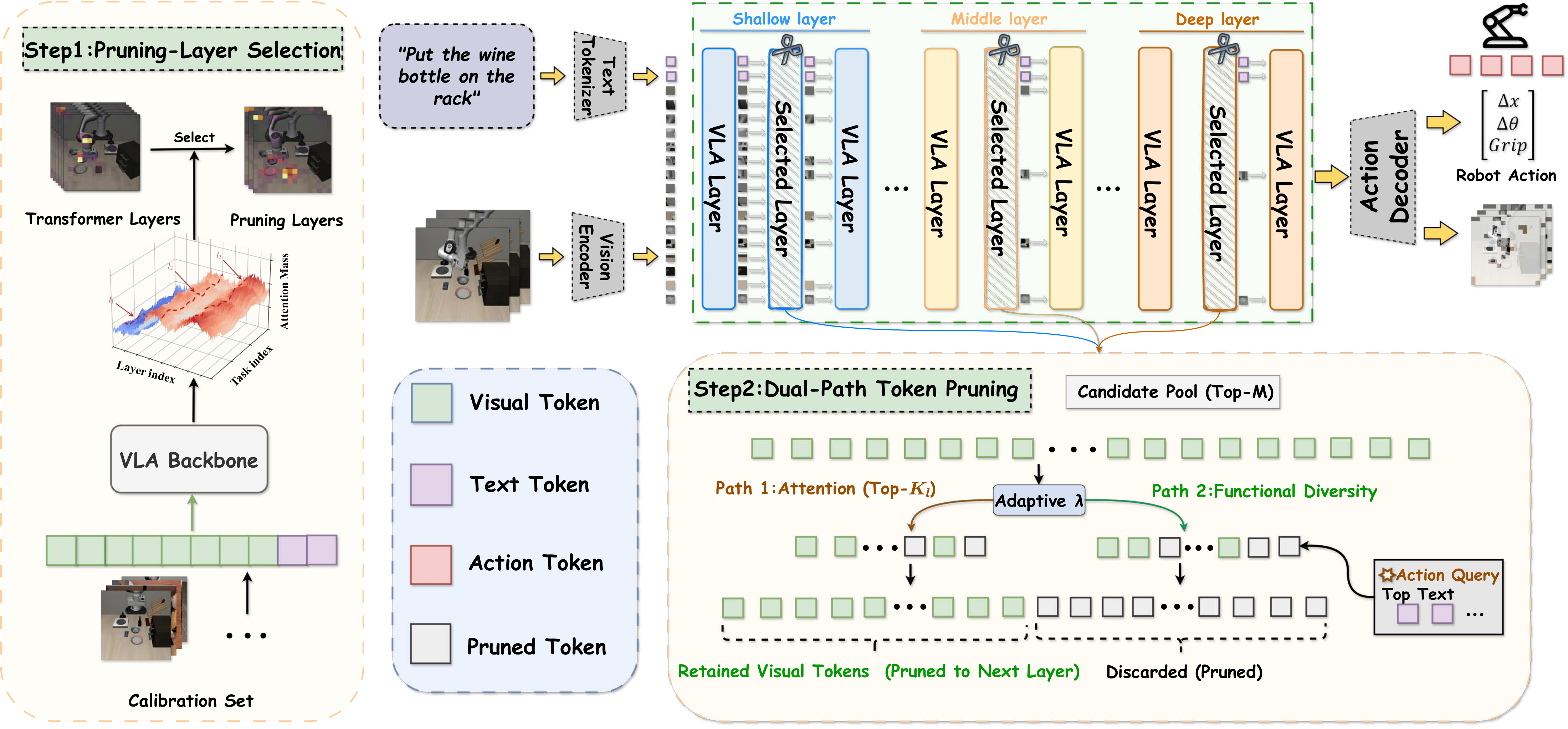}
    \caption{Overview of the framework.}
    \label{fig:overview}
\end{figure}

Figure~\ref{fig:overview} gives an overview of \textbf{SAPrune}. SAPrune uses layer-wise action-to-visual attention statistics to decide where pruning is applied. It first selects pruning layers from rollout-level action-attention dynamics, and then applies a layer-local dual-path pruning rule. 

\subsection{Pruning-layer selection from action-attention dynamics}
\label{sec:layer_selection}

\begin{figure*}[t]
    \centering

    \begin{subfigure}[t]{0.32\textwidth}
        \centering
        \includegraphics[width=\textwidth]{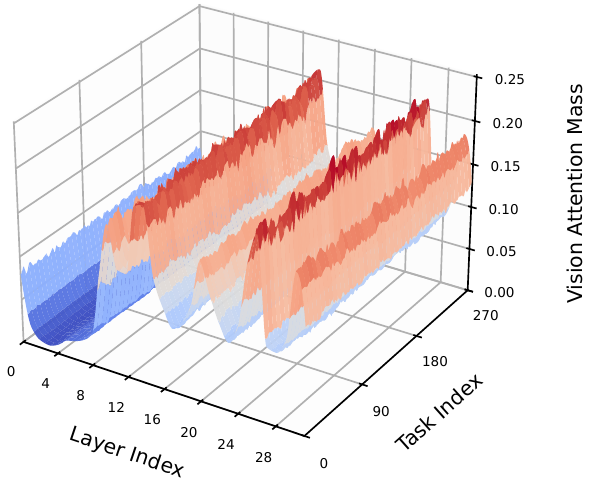}
        \caption{OpenVLA}
        \label{fig:openvla_3d_attention}
    \end{subfigure}
    \hfill
    \begin{subfigure}[t]{0.32\textwidth}
        \centering
        \includegraphics[width=\textwidth]{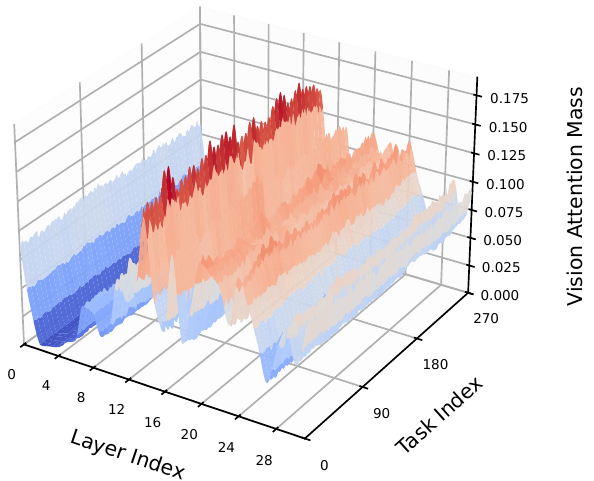}
        \caption{OpenVLA-OFT}
        \label{fig:openvla_oft_3d_attention}
    \end{subfigure}
    \hfill
    \begin{subfigure}[t]{0.32\textwidth}
        \centering
        \includegraphics[width=\textwidth]{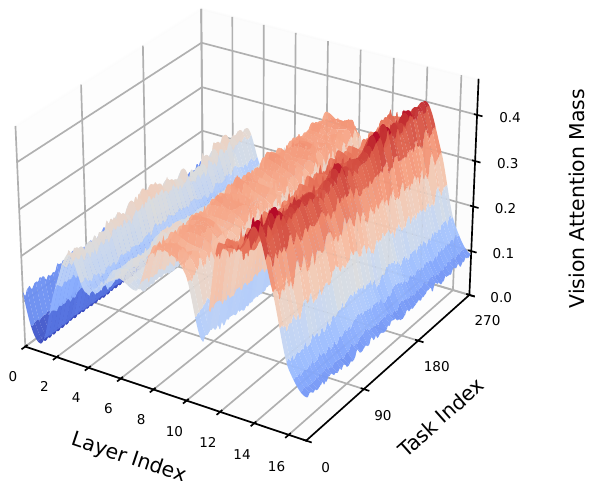}
        \caption{$\pi_0$}
        \label{fig:pi0_3d_attention}
    \end{subfigure}

    \caption{
    Task-level variation of layer-wise action-to-visual attention mass.
    The x-axis denotes the layer index, the y-axis denotes the calibration task index, and the z-axis denotes the vision attention mass.
    }
    \label{fig:task_level_attention_3d}
\end{figure*}

Given a VLA backbone, we select pruning layers by detecting stable stages in
rollout-level action-to-visual attention dynamics. We measure the layer-wise
top-\(\kappa\) action-to-visual attention mass and its inter-layer redistribution,
and then choose stable layers from the shallow, middle, and deep regions.
Calibration only logs attention statistics and uses no rewards, success labels,
or task outcomes. By default, we use LIBERO~\citep{liu2023libero} for
calibration; we also consider SIMPLER, real-world, and mixed simulation-real
splits with 30, 40, and 270 rollouts, respectively. Figure~\ref{fig:task_level_attention_3d}
visualizes task-level averages and shows that the depth-wise attention trend is
largely consistent across calibration tasks. This visualization is used only for
stability analysis, while pruning-layer selection uses the rollout-averaged
statistic defined below.

Using the action-to-visual attention defined in Sec.~\ref{sec:vla_prelim}, we
compute the top-\(\kappa\) attention mass at each layer and rollout step:
\begin{equation}
S_l(t)=\sum_{i\in \mathcal{T}_\kappa^{(l,t)}} a_i^{(l,t)},
\qquad
\mu_l=\frac{1}{T}\sum_{t=1}^{T}S_l(t),
\label{eq:layerwise_attention_mass}
\end{equation}
where \(\mathcal{T}_\kappa^{(l,t)}\) contains the indices of the top
\(\lfloor \kappa N \rfloor\) visual tokens ranked by
\(\mathbf{a}^{(l,t)}\), \(N\) is the number of visual tokens, \(t\) indexes all
calibration rollout steps, and \(\mu_l\) is the rollout-averaged top-\(\kappa\)
attention mass used for pruning-layer selection. Here, \(\kappa\) denotes the
fraction of visual tokens used to compute the top-\(\kappa\) attention mass. We
use \(\kappa=0.125\) by default. For Figure~\ref{fig:task_level_attention_3d},
we group rollout steps by calibration task \(n\) and plot the task-averaged
mass \(S_l^{(n)}=|\mathcal{D}_n|^{-1}\sum_{t\in\mathcal{D}_n}S_l(t)\), where
\(\mathcal{D}_n\) is the set of rollout steps from task \(n\).

To make the rule consistent across backbones of different depths, we partition
candidate pruning layers into three contiguous relative-depth regions:
\(\mathcal{R}_1\), \(\mathcal{R}_2\), and \(\mathcal{R}_3\), corresponding to
shallow, middle, and deep layers. The last Transformer layer is excluded from
candidate pruning locations because pruning there would not reduce subsequent
computation. We select one pruning layer from each region, denoted by
\(l_1\in\mathcal{R}_1\), \(l_2\in\mathcal{R}_2\), and
\(l_3\in\mathcal{R}_3\).

To suppress single-layer noise, we smooth \(\{\mu_l\}\) by a three-point moving
average and obtain \(\{\tilde{\mu}_l\}\). We then compute the inter-layer shift
and two quantile-based thresholds:
\begin{equation}
d_l = |\tilde{\mu}_{l+1}-\tilde{\mu}_l|,
\qquad
\tau_s = Q_{q_s}(\{d_l\}_{l=0}^{L-2}),
\qquad
\tau_r = Q_{q_r}(\{d_l\}_{l=0}^{L-2}).
\label{eq:shift_threshold}
\end{equation}
Here, \(d_l\) measures the attention shift between adjacent layers. We use
\(\tau_s\) for plateau detection and \(\tau_r\) for redistribution-event
detection. A layer is regarded as a plateau start if the following \(m\)
inter-layer shifts are all below \(\tau_s\); the set of plateau starts is
denoted by \(\mathcal{P}\). A redistribution event is a maximal consecutive
segment of indices for which \(d_l>\tau_r\).

For the middle and deep regions, we identify the dominant redistribution event.
Let \(\mathcal{E}_r\) denote the set of redistribution events within
\(\mathcal{R}_r\), where \(r\in\{2,3\}\). When \(\mathcal{E}_r\) is non-empty,
we define
\begin{equation}
[a_r^\star,b_r^\star]
=
\operatorname*{arg\,max}_{[a,b]\in\mathcal{E}_r}
\sum_{u=a}^{b} d_u,
\qquad r\in\{2,3\}.
\label{eq:dominant_redistribution_event}
\end{equation}
Here, \([a_r^\star,b_r^\star]\) is the redistribution interval with the largest
cumulative attention shift in region \(\mathcal{R}_r\).

We define the valid search regions as
\begin{equation}
\bar{\mathcal{R}}_r =
\begin{cases}
\{p\in\mathcal{R}_1\mid p\ge l_{\min}\}, 
& r=1,\ \{p\in\mathcal{R}_1\mid p\ge l_{\min}\}\neq\varnothing,\\[2pt]
\mathcal{R}_1,
& r=1,\ \{p\in\mathcal{R}_1\mid p\ge l_{\min}\}=\varnothing,\\[2pt]
\{p\in\mathcal{R}_r\mid p>b_r^\star\}, 
& r\in\{2,3\},\ \mathcal{E}_r\neq\varnothing,\ 
\{p\in\mathcal{R}_r\mid p>b_r^\star\}\neq\varnothing,\\[2pt]
\mathcal{R}_r, 
& r\in\{2,3\}\ \text{otherwise}.
\end{cases}
\label{eq:valid_search_regions}
\end{equation}
Here, \(l_{\min}\) is the earliest allowed pruning layer. For the middle and
deep regions, we search after the dominant redistribution interval when it
exists; otherwise, we fall back to the full region.

Within each valid search region, we choose the earliest plateau start if
available:
\begin{equation}
\resizebox{0.96\columnwidth}{!}{$\displaystyle
l_r=
\begin{cases}
\min\left(\mathcal{P}\cap\bar{\mathcal{R}}_r\right), 
& \text{if } \mathcal{P}\cap\bar{\mathcal{R}}_r\neq\varnothing,\\[4pt]
\operatorname*{arg\,min}_{p\in\bar{\mathcal{R}}_r}
\frac{
\sum_{u=p}^{\min(p+m-1,L-2)}d_u
}{
\min(p+m-1,L-2)-p+1
},
& \text{otherwise},
\end{cases}
\quad r\in\{1,2,3\}.
$}
\label{eq:selected_pruning_layers}
\end{equation}
In the fallback case, ties are resolved by choosing the earliest layer. Thus,
\(l_r\) is either the earliest plateau start in the valid region or the layer
with the smallest local average shift. In practice, this deterministic selection
also makes the offline calibration outcome reproducible across runs with the
same recorded attention statistics. The complete procedure is summarized in
Algorithm~\ref{alg:offline_layer_selection}. Unless otherwise stated, we use
\(q_s=0.4\), \(q_r=0.8\), \(m=2\), and \(l_{\min}=1\).

\subsection{Layer-Local Dual-Path Pruning}
\label{sec:dual_path_pruning}

At each selected pruning layer \(l\), the module receives the current visual token set \(V^{(l)}=\{z_i^{(l)}\}_{i=1}^{N_l}\) and keeps \(K_l\) tokens. Let \(q_a^{(l)}\) denote the action query and \(X^{(l)}=\{x_1^{(l)},\ldots,x_M^{(l)}\}\) denote the text-token representations. A simple baseline is to keep the top-\(K_l\) tokens ranked by action attention. However, this rule often collapses the retained set onto a small highly attended region and may remove nearby objects, spatial anchors, or background cues needed for control. SAPrune therefore splits token retention into two paths. The attention path keeps the highest-attention tokens, while the coverage path selects additional candidates whose query-response patterns are not already represented by the retained set.

\paragraph{Candidate pool and reliability-adaptive budget.}
We first restrict token selection to an expanded action-guided candidate pool and
construct an attention-ranked probe set:
\begin{equation}
C_l=\mathrm{TopM}\!\left(a^{(l)},\lfloor \gamma K_l \rfloor\right),
\qquad
A_l^{\mathrm{probe}}=
\mathrm{TopK}\!\left(a^{(l)},\lfloor \rho K_l \rfloor\right),
\label{eq:candidate_probe}
\end{equation}
where $a_i^{(l)}$ is the attention from the action query to visual token $i$.
The candidate pool \(C_l\) is larger than the final budget, allowing the second path to search beyond the top-\(K_l\) tokens while still avoiding the full visual sequence. The probe set \(A_l^{probe}\) is used to test whether the high-attention tokens already span the main candidate subspace.

To determine the budget split, we compute two reliability signals. First, from the
attention-ranked probe tokens, we construct an orthonormal basis $U_l$ using thin QR
decomposition over their normalized features. For each candidate $j\in C_l$, let
$w_j^{(l)}$ be its action-attention weight normalized within $C_l$. We measure how well
the probe basis explains the weighted candidate space by
\begin{equation}
R_l =
\frac{
\sum_{j \in C_l} w_j^{(l)}
\left\|
z_j^{(l)} - U_l U_l^\top z_j^{(l)}
\right\|_2^2
}{
\sum_{j \in C_l} w_j^{(l)}
\left\| z_j^{(l)} \right\|_2^2
},
\qquad
P_l=1-R_l.
\label{eq:preservation_ratio}
\end{equation}
Here, $P_l$ is a preservation score. A larger $P_l$ means that the attention-ranked
probe tokens already capture more of the dominant action-relevant candidate structure.

Second, we estimate how concentrated the current action attention is within the
candidate pool using normalized entropy:
\begin{equation}
H_l=-\sum_{j\in C_l}w_j^{(l)}\log w_j^{(l)},
\qquad
A_l=1-\frac{H_l}{\log |C_l|}.
\label{eq:attention_concentration}
\end{equation}
A larger \(A_l\) indicates that attention is concentrated on a few candidates, while a smaller value suggests a more diffuse distribution and therefore leaves more room for coverage-based selection.

The preservation score $P_l$ and the concentration score $A_l$ jointly determine the
budget allocation:
\begin{equation}
\eta_l=\sqrt{P_lA_l},
\qquad
\lambda_l=(1-\eta_l)\lambda_{\min}+\eta_l\lambda_{\max},
\qquad
K_l^{\mathrm{attn}}=\lfloor \lambda_lK_l\rfloor,
\quad
K_l^{\mathrm{cov}}=K_l-K_l^{\mathrm{attn}}.
\label{eq:budget_split}
\end{equation}
Thus, SAPrune allocates more tokens to the attention path when the high-attention candidates are both representative and concentrated. Otherwise, it assigns more tokens to the coverage path.

\paragraph{Attention-core path.}
The attention path keeps the top-\(K_l^{attn}\) tokens in \(C_l\) ranked by action attention, forming \(S_l^{attn}\). These tokens provide the high-confidence subset that the model currently queries most strongly.

\paragraph{Functional-diversity path.}
The coverage path fills the remaining budget with tokens that are not necessarily top-ranked by action attention but respond differently to the action and instruction queries. Since text tokens are kept unpruned, we use them to construct instruction-conditioned queries. Specifically, we rank non-special text tokens by their head-averaged attention from the action query and keep
the top-\(K_x\) tokens to form \(Q_x^{(l)}\). The query family used by the coverage path is
\begin{equation}
Q_l=\{q_a^{(l)}\}\cup Q_x^{(l)}.
\label{eq:query_set}
\end{equation}
The action query captures the control state, while the selected text queries provide instruction-specific context for measuring token coverage.

For each candidate visual token $i\in C_l$, we define a query-conditioned functional
signature:
\begin{equation}
r_i^{(l)}
=
\big[A(q,i)\big]_{q\in Q_l}\in\mathbb{R}^{|Q_l|},
\label{eq:functional_signature}
\end{equation}
where $A(q,i)$ denotes the head-averaged attention from query $q$ to visual token $i$.
Unlike feature-space diversity, this signature measures how a visual token functionally
responds to the current action-language query family. Therefore, two tokens are regarded
as redundant if they provide similar query-response patterns, even if their hidden-state
features are not identical.

We initialize the retained set as $R=S_l^{\mathrm{attn}}$. After $\ell_2$ normalizing
the functional signatures, we aggregate the total functional evidence in the candidate
pool as $r_{\mathrm{full}}^{(l)}$ and the evidence already covered by $R$ as
$r_R^{(l)}$. Their difference defines the residual evidence
$m_R^{(l)}=r_{\mathrm{full}}^{(l)}-r_R^{(l)}$, which represents the portion of
action-language evidence not yet covered by the retained tokens. The coverage path then
greedily selects the candidate that best explains this residual while avoiding redundancy:
\begin{equation}
i^\star
=
\arg\max_{i\in C_l\setminus R}
\left[
\left\langle \hat r_i^{(l)},\hat m_R^{(l)}\right\rangle
-
\beta\max_{j\in R}
\cos\!\left(r_i^{(l)},r_j^{(l)}\right)
\right].
\label{eq:coverage_selection}
\end{equation}
The first term favors tokens aligned with the missing functional evidence, while the
second term suppresses tokens redundant with the already retained set. After each
selection, the chosen token is added to both $S_l^{\mathrm{cov}}$ and $R$, and the
process continues until $K_l^{\mathrm{cov}}$ coverage tokens are selected.

Finally, the retained visual tokens are obtained by merging the two paths:
$S_l=S_l^{\mathrm{attn}}\cup S_l^{\mathrm{cov}}$. If duplicate removal leads to
underfilling, the remaining slots are refilled from unused candidates according to
action-attention rank until $|S_l|=K_l$. The retained set $S_l$ is then forwarded to
subsequent Transformer layers. Unless otherwise specified, we set \(\gamma = 1.75\), \(\rho = 0.8\), \(\lambda_{\min}=0.1\), \(\lambda_{\max}=0.9\), \(K_x=6\), and \(\beta=0.3\).

\begin{table*}[t]
\centering
\small
\setlength{\tabcolsep}{4.0pt}
\renewcommand{\arraystretch}{1.12}
\caption{\textbf{Comparison under reported/default token settings on LIBERO~\citep{liu2023libero}.}
Red bold marks the best result among pruned methods within each backbone.}
\label{tab:main_libero}

\resizebox{\textwidth}{!}{%
\begin{tabular}{lccccccccc}
\toprule
\multirow{2}{*}{Method} 
& \multirow{2}{*}{Tokens} 
& \multicolumn{5}{c}{LIBERO Success Rate (\%, $\uparrow$)} 
& \multirow{2}{*}{\makecell{Latency\\(ms, $\downarrow$)}} 
& \multirow{2}{*}{Speedup ($\uparrow$)} 
& \multirow{2}{*}{\makecell{FLOPs Ratio\\(\%, $\downarrow$)}} \\
\cmidrule(lr){3-7}
& & Spatial & Object & Goal & Long & Avg. & & & \\
\midrule

\rowcolor{sectiongray}
\multicolumn{10}{c}{\textbf{OpenVLA~\citep{kim2025openvla}}} \\
\midrule
OpenVLA~\citep{kim2025openvla}           
& 256 & 85.3 & 85.9 & 76.1 & 52.3 & 74.9 & 178.46 & 1.000$\times$ & 100.0 \\
+ FastV~\citep{chen2024fastv}            
& 128 & 81.7 & 80.5 & 73.1 & 47.9 & 70.8 & 139.88 & 1.276$\times$ & 59.2 \\
+ DivPrune~\citep{alvar2025divprune}         
& 128 & 81.1 & 78.8 & 71.7 & 47.6 & 69.8 & 136.10 & 1.311$\times$ & 62.3 \\
+ SP-VLA~\citep{li2025spvla}           
& 128 & 83.5 & 82.6 & 74.1 & 48.7 & 72.2 & 127.07 & 1.404$\times$ & 68.6 \\
+ VLA-ADP~\citep{pei2025vlaadp}          
& 76  & 84.2 & \best{85.2} & 74.9 & 50.3 & 73.7 & 130.18 & 1.371$\times$ & 45.7 \\
+ EfficientVLA(T-only)~\citep{yang2025efficientvla} 
& 58  & 80.7 & 80.6 & 70.0 & 46.4 & 69.4 & 115.64 & 1.543$\times$ & 42.1 \\
\rowcolor{oursbg}
+ \textbf{SAPrune}    
& 32  & \best{84.9} & 85.1 & \best{75.3} & \best{50.8} & \best{74.0} & \best{110.96} & \best{1.608$\times$} & \best{39.5} \\

\midrule
\rowcolor{sectiongray}
\multicolumn{10}{c}{\textbf{OpenVLA-OFT~\citep{kim2025openvlaoft}}} \\
\midrule
OpenVLA-OFT~\citep{kim2025openvlaoft}           
& 256 & 95.5 & 98.8 & 96.6 & 90.7 & 95.4 & 103.59 & 1.000$\times$ & 100.0 \\
+ FastV~\citep{chen2024fastv}            
& 128 & 94.3 & 96.5 & 92.8 & 87.1 & 92.7 & 72.42 & 1.430$\times$ & 59.9 \\
+ DivPrune~\citep{alvar2025divprune}         
& 128 & 91.8 & 92.1 & 89.6 & 84.1 & 89.4 & 73.83 & 1.403$\times$ & 58.5 \\
+ SP-VLA~\citep{li2025spvla}           
& 128 & 93.8 & 96.5 & 94.4 & 85.1 & 92.5 & 68.85 & 1.505$\times$ & 61.7 \\
+ VLA-ADP~\citep{pei2025vlaadp}          
& 76  & 94.5 & \best{98.3} & 95.3 & \best{90.0} & 94.5 & 71.11 & 1.457$\times$ & 46.3 \\
+ EfficientVLA(T-only)~\citep{yang2025efficientvla} 
& 58  & 92.0 & 92.7 & 91.0 & 84.1 & 90.0 & 63.89 & 1.621$\times$ & 41.0 \\
\rowcolor{oursbg}
+ \textbf{SAPrune}    
& 32  & \best{95.2} & 98.1 & \best{95.9} & 88.9 & \best{94.5} & \best{60.30} & \best{1.718$\times$} & \best{37.8} \\

\midrule
\rowcolor{sectiongray}
\multicolumn{10}{c}{\bfseries\boldmath $\bm{\pi}_{0}$~\citep{black2024pi0}} \\
\midrule
\bfseries\boldmath $\bm{\pi}_{0}$~\citep{black2024pi0}           
& 256 & 96.9 & 96.3 & 95.2 & 87.2 & 93.9 & 78.49 & 1.000$\times$ & 100.0 \\
+ FastV~\citep{chen2024fastv}            
& 128 & 93.3 & 92.5 & 89.6 & 80.0 & 88.8 & 58.19 & 1.349$\times$ & 60.0 \\
+ DivPrune~\citep{alvar2025divprune}         
& 128 & 91.0 & 92.1 & 87.2 & 79.8 & 87.5 & 60.40 & 1.300$\times$ & 55.8 \\
+ SP-VLA~\citep{li2025spvla}           
& 128 & 94.1 & 94.2 & 91.7 & 82.9 & 90.7 & 57.74 & 1.359$\times$ & 64.2 \\
+ VLA-ADP~\citep{pei2025vlaadp}          
& 76  & \best{96.3} & 95.7 & \best{94.9} & 84.7 & 92.9 & 58.60 & 1.339$\times$ & 46.7 \\
\rowcolor{oursbg}
+ \textbf{SAPrune}    
& 32  & 96.2 & \best{95.8} & 94.4 & \best{86.0} & \best{93.1} & \best{51.22} & \best{1.532$\times$} & \best{38.6} \\
\bottomrule
\end{tabular}%
}
\end{table*}

\section{Experiment}
\label{sec:experiment}
We evaluate SAPrune on two simulation benchmarks, LIBERO~\citep{liu2023libero} and SIMPLER~\citep{li2024simpler}, as well as on four real-world manipulation tasks. We compare our method with several representative efficient inference baselines, including FastV~\citep{chen2024fastv}, DivPrune~\citep{alvar2025divprune}, SP-VLA~\citep{li2025spvla}, VLA-ADP~\citep{pei2025vlaadp}, and EfficientVLA~\citep{yang2025efficientvla}, and instantiate them on three VLA backbones: OpenVLA~\citep{kim2025openvla}, OpenVLA-OFT~\citep{kim2025openvlaoft}, and $\pi_0$~\citep{black2024pi0}. For real-world evaluation, we deploy the system on an AgileX Piper. All latency results are measured on an NVIDIA RTX 5090 GPU. Unless stated otherwise, latency is measured with PyTorch SDPA for both the full model and SAPrune, without model compilation. FlashAttention-2 results are reported in Appendix~\ref{app:backend}.

\subsection{Experimental Setup}

\noindent\textbf{Compared Methods.}
We compare SAPrune with representative token-pruning baselines for efficient VLA inference, including FastV~\citep{chen2024fastv}, DivPrune~\citep{alvar2025divprune}, SP-VLA~\citep{li2025spvla}, VLA-ADP~\citep{pei2025vlaadp}, and EfficientVLA~\citep{yang2025efficientvla}. These methods cover the major pruning paradigms relevant to our setting: FastV~\citep{chen2024fastv} performs attention-based top-token selection, DivPrune~\citep{alvar2025divprune} preserves visually diverse tokens to reduce redundancy, SP-VLA~\citep{li2025spvla} applies structured pruning tailored to VLA inference, VLA-ADP~\citep{pei2025vlaadp} adaptively adjusts token retention during inference, and EfficientVLA~\citep{yang2025efficientvla} combines token relevance with feature-space diversity in a progressive pruning framework. For the main comparison, we follow the reported or default token settings of these methods rather than enforcing a unified token budget for all baselines. Specifically, the original models use 256 visual tokens, while different baselines adopt their own retained-token settings; our method keeps only 32 final visual tokens. To eliminate the effect of unequal token budgets, we further provide controlled comparisons under the same 32-token setting.
\begin{wraptable}{r}{0.50\columnwidth}
\centering
\small
\setlength{\tabcolsep}{3pt}
\renewcommand{\arraystretch}{0.9}
\caption{Comparison of SAPrune within the OpenVLA~\citep{kim2025openvla} model in the SIMPLER~\citep{li2024simpler} environment.
Red bold marks the best result among pruned methods.}
\label{tab:simpler}
\resizebox{0.48\columnwidth}{!}{%
\begin{tabular}{lccccc}
\toprule
\multirow{2}{*}{Method} & \multirow{2}{*}{Tokens} & \multicolumn{4}{c}{Success Rate (\%, $\uparrow$)} \\
\cmidrule(lr){3-6}
& & PickCoke & MoveNear & Open/Close & Avg. \\
\midrule
\rowcolor{sectiongray}
\multicolumn{6}{c}{\textbf{OpenVLA~\citep{kim2025openvla}}} \\
\midrule
OpenVLA~\citep{kim2025openvla}       & 256 & 54.7 & 57.8 & 48.1 & 53.5 \\
+ FastV~\citep{chen2024fastv}          & 32  & 19.6 & 20.1 & 16.5 & 18.7 \\
+ SP-VLA~\citep{li2025spvla}           & 32  & 35.4 & 36.1 & 35.4 & 35.6 \\
+ VLA-ADP~\citep{pei2025vlaadp}        & 32  & 39.4 & 42.5 & 38.2 & 40.0 \\
+ EfficientVLA~\citep{yang2025efficientvla} & 32  & 43.3 & 44.0 & 39.7 & 42.3 \\
\rowcolor{oursbg}
+ \textbf{SAPrune}                         & 32  & \best{53.2} & \best{56.3} & \best{47.9} & \best{52.5} \\
\bottomrule
\end{tabular}%
}
\end{wraptable}

\noindent\textbf{Evaluation Metrics.}
We report success rate, FLOPs Ratio, latency, and speedup. Success rate measures task performance. FLOPs Ratio measures the relative theoretical computation cost with respect to the original model. Unless labeled as E2E latency, latency measures the VLA model inference time in milliseconds, including pruning overhead. E2E latency measures the policy pipeline from observation preprocessing to action-command transmission, excluding physical actuation and low-level servo time. Speedup is computed as the ratio between the latency of the original model and that of the accelerated model. Lower FLOPs Ratio and latency indicate better efficiency, while higher success rate and speedup indicate better overall performance. For consistency, the quantity previously denoted as \emph{inference time} in the tables is referred to as latency throughout this paper.
\begin{table*}[!htbp]
\centering
\small
\setlength{\tabcolsep}{4pt}
\renewcommand{\arraystretch}{1.05}
\caption{Comparison on $\pi_{0.5}$ under reported/default token settings on LIBERO.
Red bold marks the best result among pruned methods.}
\label{tab:pi05}
\resizebox{\textwidth}{!}{%
\begin{tabular}{lcccccccc}
\toprule
\textbf{Method} & \textbf{Spatial SR} & \textbf{Object SR} & \textbf{Goal SR} & \textbf{Long SR} & \textbf{Avg. SR} & \textbf{Visual tokens / image} & \textbf{Retention} & \textbf{E2E latency (ms)} \\
\midrule
$\pi_{0.5}$~\citep{black2025pi05} & 97.4 & 98.4 & 97.6 & 93.0 & 96.6 & 256 & 100\% & 91.47 \\
+ FastV~\citep{chen2024fastv} & 95.8 & 95.6 & 94.1 & 89.3 & 93.7 & 128 & 50\% & 70.84 \\
+ VLA-Pruner~\citep{liu2025vlapruner} & 95.8 & 96.2 & \best{97.1} & \best{91.8} & 95.2 & 64 & 25\% & 62.73 \\
\rowcolor{oursbg}
+ \textbf{SAPrune} & \best{97.2} & \best{98.0} & 96.9 & 91.5 & \best{95.9} & 32 & 12.5\% & \best{59.68} \\
\bottomrule
\end{tabular}%
}
\end{table*}
\begin{table*}[t]
\centering
\setlength{\tabcolsep}{5pt}
\renewcommand{\arraystretch}{1.08}
\caption{\textbf{Real-world manipulation results on AgileX Piper with the $\pi_0$~\citep{black2024pi0} backbone.}
Each task is evaluated over 100 trials.
Bold marks the best result among pruned methods.}
\label{tab:realworld_results}
\resizebox{\textwidth}{!}{%
\begin{tabular}{lccccccc}
\toprule
\multirow{2}{*}{\textbf{Method}} 
& \multicolumn{5}{c}{\textbf{Success Rate $\uparrow$}} 
& \multirow{2}{*}{\textbf{Latency (ms)$\downarrow$}} 
& \multirow{2}{*}{\textbf{FLOPs Ratio (\%)$\downarrow$}} \\
\cmidrule(lr){2-6}
& \textbf{Cube Grasping} 
& \textbf{Pick\&Place} 
& \textbf{Select-Pick\&Place}
& \textbf{Multi-Cube Stacking} 
& \textbf{Average} 
& & \\
\midrule
\bfseries\boldmath $\bm{\pi}_{0}$~\citep{black2024pi0}
& 75.0
& 63.0
& 54.0
& 30.0
& 55.5
& 78.16
& 100.0 \\

+ FastV~\citep{chen2024fastv}
& 69.0
& 59.0
& 47.0
& 23.0
& 49.5
& 58.92
& 60.6 \\

+ DivPrune~\citep{alvar2025divprune}
& 70.0
& 60.0
& 47.0
& 24.0
& 50.3
& 61.38
& 64.8 \\

+ SP-VLA~\citep{li2025spvla}
& 72.0
& 61.0
& 52.0
& 25.0
& 52.5
& 58.40
& 63.4 \\

+ VLA-ADP~\citep{pei2025vlaadp}
& \textbf{74.0}
& 64.0
& \textbf{54.0}
& \textbf{26.0}
& \textbf{54.5}
& 59.30
& 47.3 \\

+ \textbf{SAPrune}
& 73.0
& \textbf{65.0}
& 53.0
& 25.0
& 54.0
& \textbf{51.45}
& \textbf{39.3} \\
\bottomrule
\end{tabular}%
}
\end{table*}
\subsection{Results on Simulation Environment}

\noindent\textbf{Results on LIBERO.}
Table~\ref{tab:main_libero} reports the main results on LIBERO~\citep{liu2023libero}. 
Across OpenVLA~\citep{kim2025openvla}, OpenVLA-OFT~\citep{kim2025openvlaoft}, and $\pi_0$~\citep{black2024pi0}, SAPrune maintains success rates close to the full-token models while using a much smaller visual-token budget. 
It also brings clear efficiency gains, achieving 1.608$\times$, 1.718$\times$, and 1.532$\times$ speedups on the three backbones, respectively, with substantial FLOPs reduction. 
Compared with existing pruning baselines, SAPrune achieves higher average success under a more aggressive final pruning ratio, indicating that aggressive pruning can remain reliable when token reduction is applied at more stable action-attention stages. 
The main degradation appears on the Long suite, where long-horizon manipulation is more sensitive to accumulated execution errors and extreme token reduction. Figure~\ref{fig:attention_dynamics_prune} provides additional LIBERO visualizations that qualitatively support the observed robustness under high-ratio pruning. We also evaluate SAPrune on $\pi_{0.5}$~\citep{black2025pi05} on LIBERO. The complete results are shown in Table~\ref{tab:pi05}. SAPrune reaches 95.9\% average success versus 96.6\% for the full model, while reducing E2E latency from 91.47 ms to 59.68 ms with 32 visual tokens versus 256 for the full model. This result extends the evaluation to a newer VLA backbone.

\noindent\textbf{Results on SIMPLER.}
Table~\ref{tab:simpler} reports the results on the visible tasks in SIMPLER~\citep{li2024simpler} under a compact 32-token budget. 
Under this aggressive pruning setting, SAPrune maintains substantially stronger task performance than other pruning baselines on the OpenVLA~\citep{kim2025openvla} backbone. 
While attention-based, structured, adaptive, and feature-diversity-based baselines suffer clear success drops after reducing the visual tokens to 32, SAPrune remains comparatively stable across PickCoke, MoveNear, and Open/Close. 
These results further indicate that, under high-ratio token reduction, selecting pruning stages from action-attention dynamics and preserving complementary action-language evidence are critical for maintaining reliable manipulation performance.

\subsection{Results on Real-World Environment}

Table~\ref{tab:realworld_results} reports the real-world manipulation results on AgileX Piper with the $\pi_0$~\citep{black2024pi0} backbone. The results show that SAPrune achieves the lowest inference cost among pruned methods while maintaining competitive task success rates. Figure~\ref{fig:real_agx} provides qualitative rollouts on the AgileX Piper robot. We also measure end-to-end latency on the same real-robot runs. The complete comparison is shown in Table~\ref{tab:e2e_latency}. SAPrune reduces latency from 88.68 ms to 61.97 ms and raises the policy-pipeline frequency from 11.28 to 16.14 Hz, a 30.1\% latency reduction. Physical actuation and low-level servo time are excluded.

\subsection{Ablation Studies}
\begin{table}[!htbp]
\centering
\tiny
\setlength{\tabcolsep}{5.2pt}
\renewcommand{\arraystretch}{1.08}
\caption{\textbf{Ablation on pruning-location strategy under OpenVLA~\citep{kim2025openvla} on LIBERO~\citep{liu2023libero}.}
Red bold marks the best result among pruning variants.}
\label{tab:ablation_location}

\resizebox{\columnwidth}{!}{%
\begin{tabular}{lcccccccc}
\toprule
\multirow{2}{*}{Method} 
& \multirow{2}{*}{\makecell{Pruning\\location}} 
& \multicolumn{5}{c}{LIBERO Success Rate (\%, $\uparrow$)} 
& \multirow{2}{*}{\makecell{Latency\\(ms, $\downarrow$)}} 
& \multirow{2}{*}{\makecell{FLOPs Ratio\\(\%, $\downarrow$)}} \\
\cmidrule(lr){3-7}
& & Spatial & Object & Goal & Long & Avg. & & \\
\midrule
OpenVLA~\citep{kim2025openvla}             
& --          
& 85.3 & 85.9 & 76.1 & 52.3 & 74.9 & 178.46 & 100.0 \\

Uniform-A            
& [1, 11, 23] 
& 78.2 & 77.1 & 67.5 & 40.2 & 65.8 & 104.40 & 34.7 \\

Uniform-B            
& [5, 19, 28] 
& 79.1 & 78.4 & 69.6 & 42.5 & 67.4 & 117.85 & 42.7 \\

Single-layer         
& [2]         
& 68.2 & 67.1 & 58.5 & 26.7 & 55.1 & \best{95.64} & \best{25.5} \\

Random three layers  
& --          
& 71.5 & 72.3 & 64.1 & 32.9 & 60.2 & 119.19 & 41.1 \\

\rowcolor{oursbg}
\textbf{SAPrune}        
& [2, 16, 26] 
& \best{84.9} & \best{85.1} & \best{75.3} & \best{50.8} & \best{74.0} & 110.96 & 39.5 \\
\bottomrule
\end{tabular}%
}
\end{table}

\noindent\textbf{Effect of pruning-location strategy.}
Table~\ref{tab:ablation_location} ablates the choice of pruning locations under OpenVLA~\citep{kim2025openvla} on LIBERO~\citep{liu2023libero}. Uniform, random, and single-layer pruning all lead to lower success rates, showing that pruning-layer selection is critical under high-ratio token reduction. In contrast, SAPrune achieves the best success rate among pruning-location variants while still providing substantial acceleration. This indicates that pruning layers should be determined by action-attention dynamics rather than manually fixed or randomly selected.

\begin{figure*}[!htbp]
    \centering

    \begin{subfigure}[t]{0.31\textwidth}
        \centering
        \includegraphics[width=\textwidth]{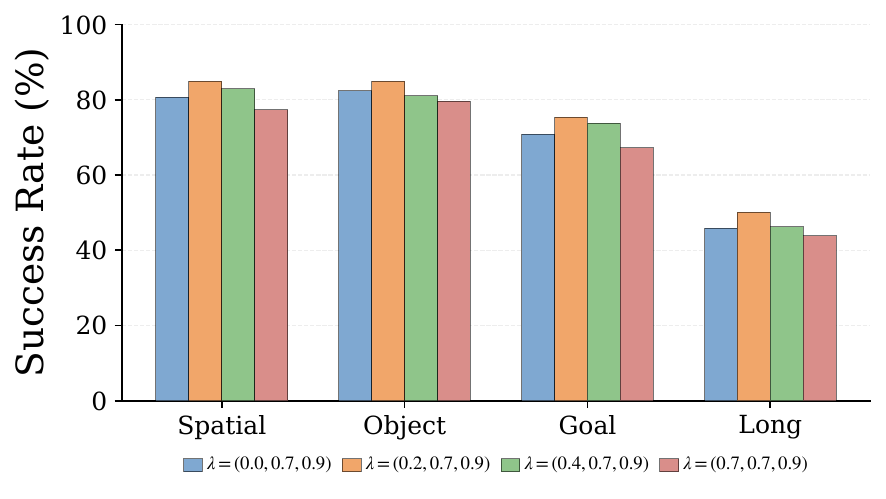}
        \caption{Ablation at Shallow Layer}
        \label{fig:ablation_shallow}
    \end{subfigure}
    \hfill
    \begin{subfigure}[t]{0.31\textwidth}
        \centering
        \includegraphics[width=\textwidth]{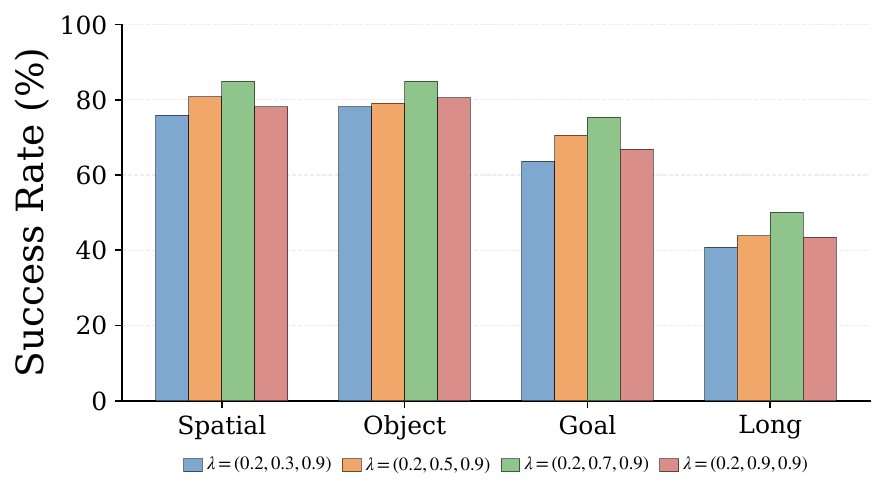}
        \caption{Ablation at Middle Layer}
        \label{fig:ablation_middle}
    \end{subfigure}
    \hfill
    \begin{subfigure}[t]{0.31\textwidth}
        \centering
        \includegraphics[width=\textwidth]{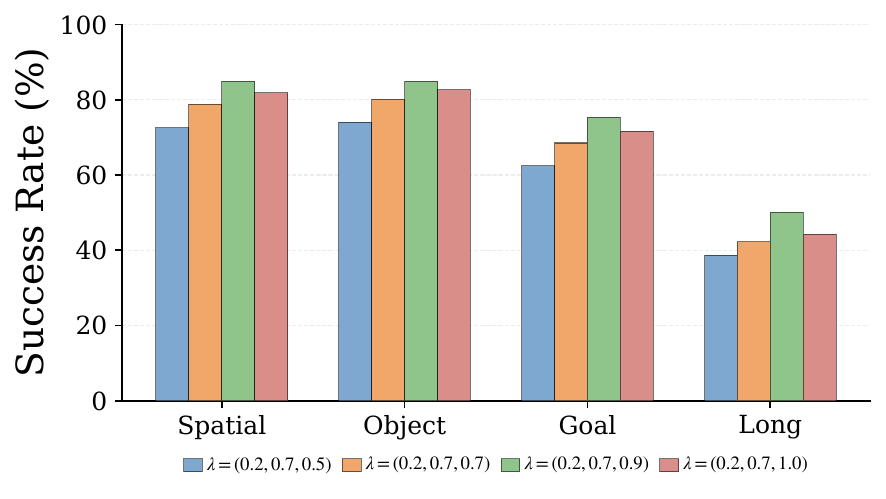}
        \caption{Ablation at Deep Layer}
        \label{fig:ablation_deep}
    \end{subfigure}

    \caption{Ablation study on balance factor.}
    \label{fig:ablation_balance_factor}
\end{figure*}

\noindent\textbf{Effect of the balance factor.}
Figure~\ref{fig:ablation_balance_factor} ablates the layer-wise balance factor $\lambda$ under OpenVLA~\citep{kim2025openvla} on LIBERO~\citep{liu2023libero}. 
The results show that the preferred allocation changes with depth. 
A small $\lambda$ works better in the shallow layer, suggesting that early pruning should keep more surrounding visual context. 
As the layer becomes deeper, larger $\lambda$ values become more effective, indicating that attention becomes a more reliable cue for selecting action-relevant tokens. 
However, the deep-layer results also show that $\lambda=1.0$ is not optimal, meaning that fully relying on attention still discards useful supporting information. 
These trends support the adaptive layer-wise allocation used in SAPrune.

\section{Conclusion}
We present SAPrune, a training-free visual token pruning framework for efficient VLA inference under aggressive token reduction. Instead of treating pruning as a static token-ranking problem, SAPrune addresses two practical questions for VLA deployment: when pruning should be applied, and which visual tokens can be safely removed. The proposed pruning-layer selection avoids arbitrary layer choices by using rollout-derived action-to-visual attention statistics to identify more reliable pruning stages. At each selected layer, the dual-path pruning rule removes redundant visual tokens while reducing the risk of pruning strongly attended action cues or useful surrounding context. This design requires no retraining or architectural modification, making SAPrune compatible with different VLA backbones.

{
\small

}

\newpage
\bibliographystyle{unsrtnat}
\bibliography{references}

\newpage
\appendix

\section{Method Details}
\subsection{Algorithm Details}
We detail SAPrune in Algorithm~\ref{alg:offline_layer_selection} and Algorithm~\ref{alg:dual_path_pruning}
\begin{algorithm}[!htbp]
\caption{Stage-Aware Pruning-Layer Selection}
\label{alg:offline_layer_selection}
\small
\begin{algorithmic}[1]
\Require Calibration action-to-visual attentions $\{a^{(l,t)}\}$,
$\kappa,q_s,q_r,m,l_{\min}$
\Ensure Pruning layers $\mathcal{L}_{\mathrm{prune}}=\{l_1,l_2,l_3\}$

\State Compute rollout-averaged top-$\kappa$ attention masses $\{\mu_l\}_{l=0}^{L-1}$
\State Smooth $\{\mu_l\}_{l=0}^{L-1}$ to obtain $\{\tilde{\mu}_l\}_{l=0}^{L-1}$
\State Compute inter-layer shifts $d_l=|\tilde{\mu}_{l+1}-\tilde{\mu}_l|$ for $l=0,\ldots,L-2$
\State $\tau_s\gets Q_{q_s}(\{d_l\}_{l=0}^{L-2})$, \quad
       $\tau_r\gets Q_{q_r}(\{d_l\}_{l=0}^{L-2})$
\State Detect plateau starts $\mathcal{P}$ using threshold $\tau_s$ and length $m$
\State Detect redistribution events $\mathcal{E}$ using threshold $\tau_r$
\State Partition candidate pruning layers into $\mathcal{R}_1,\mathcal{R}_2,\mathcal{R}_3$

\For{$r=1,2,3$}
    \If{$r=1$}
        \State $\bar{\mathcal{R}}_r\gets \{p\in\mathcal{R}_1\mid p\ge l_{\min}\}$
        \If{$\bar{\mathcal{R}}_r=\emptyset$}
            \State $\bar{\mathcal{R}}_r\gets \mathcal{R}_1$
        \EndIf
    \Else
        \State $\mathcal{E}_r\gets \{[a,b]\in\mathcal{E}\mid [a,b]\cap\mathcal{R}_r\neq\emptyset\}$
        \If{$\mathcal{E}_r\neq\emptyset$}
            \State $[a_r^\star,b_r^\star]\gets
            \arg\max_{[a,b]\in\mathcal{E}_r}\sum_{u=a}^{b}d_u$
            \State $\bar{\mathcal{R}}_r\gets \{p\in\mathcal{R}_r\mid p>b_r^\star\}$
            \If{$\bar{\mathcal{R}}_r=\emptyset$}
                \State $\bar{\mathcal{R}}_r\gets \mathcal{R}_r$
            \EndIf
        \Else
            \State $\bar{\mathcal{R}}_r\gets \mathcal{R}_r$
        \EndIf
    \EndIf
\EndFor

\For{$r=1,2,3$}
    \If{$\mathcal{P}\cap\bar{\mathcal{R}}_r\neq\emptyset$}
        \State $l_r\gets \min(\mathcal{P}\cap\bar{\mathcal{R}}_r)$
    \Else
        \State $l_r\gets \arg\min_{p\in\bar{\mathcal{R}}_r}
        \frac{
        \sum_{u=p}^{\min(p+m-1,L-2)}d_u
        }{
        \min(p+m-1,L-2)-p+1
        }$
    \EndIf
\EndFor

\State \Return $\mathcal{L}_{\mathrm{prune}}=\{l_1,l_2,l_3\}$
\end{algorithmic}
\end{algorithm}

\begin{algorithm}[t]
\caption{Layer-Local Dual-Path Visual Token Pruning}
\label{alg:dual_path_pruning}
\small
\begin{algorithmic}[1]
\Require Visual tokens $V_t^{(l)}$, text states $X^{(l)}$, action query $q_a^{(l)}$,
action attention $a^{(l)}$, keep budget $K_l$
\Ensure Retained visual-token set $S_l$

\State Construct candidate pool 
$C_l\gets \mathrm{TopM}(a^{(l)},\lfloor\gamma K_l\rfloor)$
\State Construct probe set 
$A_l^{\mathrm{probe}}\gets \mathrm{TopK}(a^{(l)},\lfloor\rho K_l\rfloor)$
\State Compute preservation score $P_l$ from the probe subspace
\State Compute attention-concentration score $A_l$ over $C_l$
\State $\eta_l\gets \sqrt{P_lA_l}$
\State $\lambda_l\gets (1-\eta_l)\lambda_{\min}+\eta_l\lambda_{\max}$
\State $K_l^{\mathrm{attn}}\gets \lfloor\lambda_lK_l\rfloor$, \quad
$K_l^{\mathrm{cov}}\gets K_l-K_l^{\mathrm{attn}}$

\Statex \textbf{Attention-core path}
\State $S_l^{\mathrm{attn}}\gets 
\mathrm{TopK}_{C_l}(a^{(l)},K_l^{\mathrm{attn}})$
\State $R\gets S_l^{\mathrm{attn}}$

\Statex \textbf{Functional-diversity path}
\State Select top text queries $Q_x^{(l)}$ by action-to-text attention
\State $Q_l\gets \{q_a^{(l)}\}\cup Q_x^{(l)}$
\State Compute functional signatures $\{r_i^{(l)}\}_{i\in C_l}$ using $Q_l$
\State $S_l^{\mathrm{cov}}\gets\emptyset$

\While{$|S_l^{\mathrm{cov}}|<K_l^{\mathrm{cov}}$ and $C_l\setminus R\neq\emptyset$}
    \State Compute residual functional evidence $m_R^{(l)}$
    \State Select
    \[
    i^\star\gets
    \arg\max_{i\in C_l\setminus R}
    \left[
    \langle \hat r_i^{(l)},\hat m_R^{(l)}\rangle
    -
    \beta\max_{j\in R}\cos(r_i^{(l)},r_j^{(l)})
    \right]
    \]
    \State $S_l^{\mathrm{cov}}\gets S_l^{\mathrm{cov}}\cup\{i^\star\}$
    \State $R\gets R\cup\{i^\star\}$
\EndWhile

\State $S_l\gets S_l^{\mathrm{attn}}\cup S_l^{\mathrm{cov}}$
\If{$|S_l|<K_l$}
    \State Refill remaining slots from unused candidates by action-attention rank
\EndIf
\State \Return $S_l$
\end{algorithmic}
\end{algorithm}
\subsection{Complexity Analysis}
\label{app:complexity}

We first estimate the multi-head attention and FFN FLOPs in the Transformer
backbone. The equations below describe this backbone cost; the reported total
FLOPs also include routing and token selection.

Let \(L\) be the number of Transformer layers, \(N_0\) the original number of
visual tokens, and \(M\) the number of non-visual tokens. The initial sequence
length is
\begin{equation}
\mu_0 = M + N_0 .
\end{equation}
The selected pruning layers in the main text use 0-based indices
\(\tilde l_1,\tilde l_2,\tilde l_3\). For FLOPs accounting, we convert them to
1-based layer indices:
\begin{equation}
l_i = \tilde l_i + 1,\qquad i\in\{1,2,3\},
\end{equation}
with \(1\le l_1<l_2<l_3<L\). Let \(K_{l_1},K_{l_2},K_{l_3}\) denote the
retained numbers of visual tokens after the three pruning stages, and define
\begin{equation}
\mu_1=M+K_{l_1},\qquad
\mu_2=M+K_{l_2},\qquad
\mu_3=M+K_{l_3}.
\end{equation}

For a Transformer layer with sequence length \(n\), hidden size \(d\), and FFN
intermediate size \(d_{\mathrm{ff}}\), we approximate its FLOPs as
\begin{equation}
C(n)=4nd^2+2n^2d+2nd\,d_{\mathrm{ff}}.
\end{equation}
Without pruning, the total backbone FLOPs are
\begin{equation}
\mathrm{FLOPs}_{\mathrm{full}}=L\,C(\mu_0).
\end{equation}

In SAPrune, pruning is applied after the selected layer outputs. Therefore, the
pruning at layer \(l_i\) affects only subsequent layers. The first \(l_1\)
layers use \(\mu_0\), layers \(l_1+1\) to \(l_2\) use \(\mu_1\), layers
\(l_2+1\) to \(l_3\) use \(\mu_2\), and layers \(l_3+1\) to \(L\) use
\(\mu_3\). Thus,
\begin{equation}
\mathrm{FLOPs}_{\mathrm{prune}}
=
l_1 C(\mu_0)
+
(l_2-l_1)C(\mu_1)
+
(l_3-l_2)C(\mu_2)
+
(L-l_3)C(\mu_3).
\end{equation}
The theoretical FLOPs ratio is
\begin{equation}
r_{\mathrm{FLOPs}}
=
\frac{\mathrm{FLOPs}_{\mathrm{prune}}}
{\mathrm{FLOPs}_{\mathrm{full}}}.
\end{equation}

In the main setting, SAPrune uses stage-wise 50\% pruning:
\begin{equation}
K_{l_1}=128,\qquad
K_{l_2}=64,\qquad
K_{l_3}=32,
\end{equation}
starting from \(N_0=256\) visual tokens. The final retained-token ratio is
therefore \(32/256=12.5\%\), corresponding to an overall pruning ratio of
\(87.5\%\). Token selection is performed only at the three pruning layers
and over restricted candidate sets.

The reported inference latency includes the selected action- and text-query
attention computation, routing, token selection, indexing, and gather operations.
On OpenVLA, SAPrune uses 39.5\% of the full-model FLOPs, including pruning costs,
while latency falls from 178.46\,ms to 110.96\,ms. Thus, a 60.5\% FLOPs reduction
gives a 37.8\% latency reduction (1.608$\times$ speedup). Memory access, kernel
launches, and other runtime costs do not decrease in direct proportion to FLOPs.
The attention-backend comparison and real-robot policy-pipeline breakdown are
reported in Appendix~\ref{app:backend} and Table~\ref{tab:e2e_latency}, respectively.
\section{Experimental Setup}
\label{sec:exp_setup}

\subsection{VLA Model Details}
\label{app:vla_backbones}

We evaluate SAPrune on representative VLA backbones, including OpenVLA, OpenVLA-OFT, $\pi_0$, and $\pi_{0.5}$. These models cover both autoregressive action-token policies and flow-based continuous-action policies, allowing us to examine whether the proposed pruning strategy generalizes across different action-generation mechanisms.

\paragraph{OpenVLA.}
OpenVLA~\citep{kim2025openvla} is a 7B open-source vision-language-action model adapted from a pretrained vision-language model for robotic control. It encodes visual observations and language instructions, and predicts discretized robot actions through autoregressive token generation. We use OpenVLA as the representative token-based VLA backbone.

\paragraph{OpenVLA-OFT.}
OpenVLA-OFT~\citep{kim2025openvlaoft} is an optimized fine-tuning variant of OpenVLA. It improves the original action-generation recipe with parallel decoding, action chunking, continuous action representation, and regression-based training objectives. Compared with the original OpenVLA, it provides stronger control performance and higher inference efficiency while preserving the same general VLA formulation. We use OpenVLA-OFT as a stronger OpenVLA-family backbone.

\paragraph{$\pi_0$ Model.}
$\pi_0$~\citep{black2024pi0} is a flow-based vision-language-action model for general robot control. Instead of decoding discretized action tokens autoregressively, it augments a pretrained vision-language model with a continuous-action flow-matching module. This makes $\pi_0$ representative of continuous-action VLA policies and complements the OpenVLA-family models in our evaluation.

\paragraph{$\pi_{0.5}$ Model.}
$\pi_{0.5}$~\citep{black2025pi05} is a newer VLA model designed for open-world generalization. We evaluate it on LIBERO and LIBERO-Plus using the same calibration and pruning procedure, without manual pruning-layer tuning.

\subsection{Acceleration Baseline Details}
\label{app:accelerated_baselines}

We compare SAPrune with representative training-free or plug-and-play acceleration baselines, including FastV, DivPrune, SP-VLA, VLA-ADP, and EfficientVLA. These methods cover attention-based pruning, diversity-based pruning, structured VLA pruning, adaptive pruning, and progressive VLA acceleration.

\paragraph{FastV.}
FastV~\citep{chen2024fastv} is a training-free attention-based pruning method for large vision-language models. It ranks visual tokens by attention-derived importance and removes low-score tokens after a selected layer. Since it relies mainly on attention relevance, we use it as an attention-only pruning baseline.

\paragraph{DivPrune.}
DivPrune~\citep{alvar2025divprune} selects visual tokens by encouraging feature-space diversity among the retained set. It reduces redundancy by preserving visually dissimilar tokens, but its diversity criterion is not explicitly conditioned on action generation. We use it as a representative diversity-based pruning baseline.

\paragraph{SP-VLA.}
SP-VLA~\citep{li2025spvla} is a VLA-oriented acceleration method that combines model scheduling with token pruning. Compared with generic VLM pruning methods, it is more tailored to VLA inference. We include it as a structured VLA pruning baseline.

\paragraph{VLA-ADP.}
VLA-ADP~\citep{pei2025vlaadp} performs action-aware dynamic pruning for efficient VLA manipulation. It adaptively adjusts token retention during inference rather than using a completely fixed pruning rule. We use it as a representative adaptive VLA pruning baseline.

\paragraph{EfficientVLA.}
EfficientVLA~\citep{yang2025efficientvla} is a recent training-free acceleration framework for VLA models. Its pruning component combines token relevance with feature-space diversity in a progressive pruning pipeline. We include EfficientVLA, especially its text-guided variant EfficientVLA(T-only), as a strong relevance-plus-diversity baseline.

\paragraph{Evaluation protocol.}
We report two groups of comparisons. In the default-budget setting, each baseline is evaluated under its reported or recommended token budget, while the original models use 256 visual tokens and SAPrune keeps 32 visual tokens. This setting reflects the practical configuration of each method. In the controlled-budget setting, we fix the final retained-token budget to 32 for all applicable pruning methods, so that performance differences mainly reflect the token-selection strategy rather than the number of remaining visual tokens.

\begin{figure}[t]
    \centering
    \includegraphics[width=\linewidth]{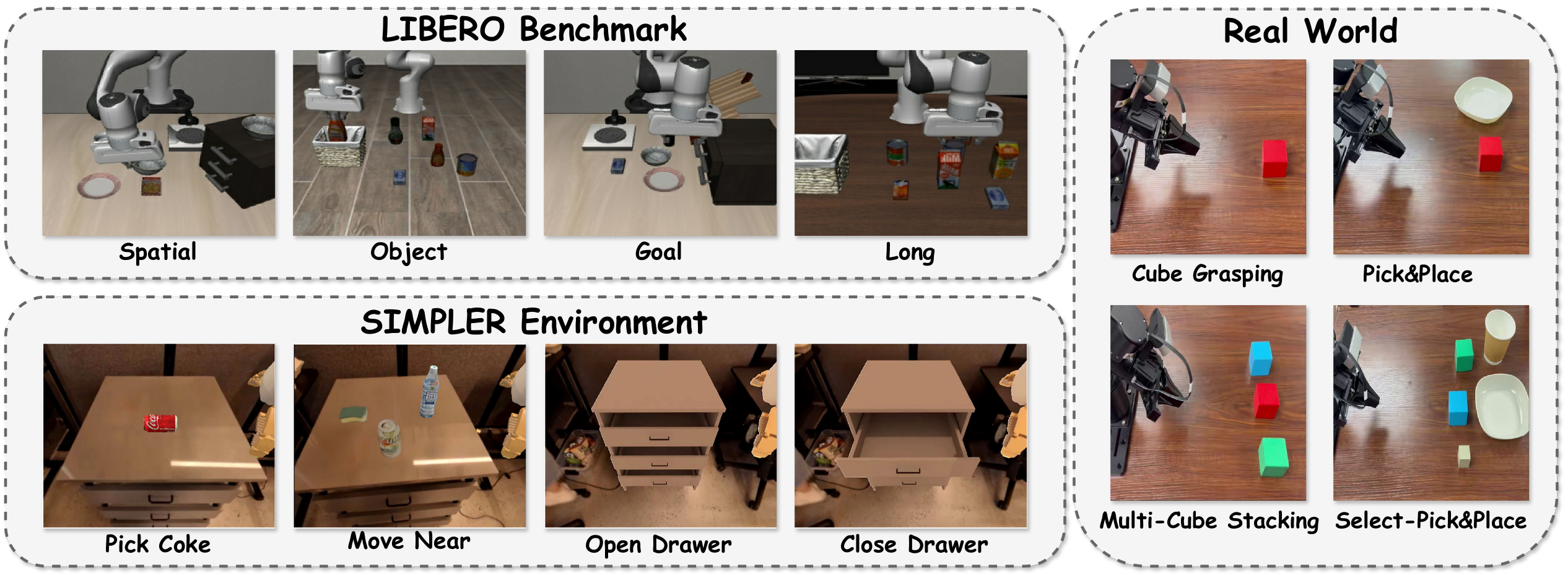}
    \caption{Tasks on LIBERO Benchmark, SIMPLER and Real World Environment.}
    \label{fig:Benchmark}
\end{figure}

\subsection{Benchmarks and Tasks}
\label{app:benchmarks}

We evaluate SAPrune on three groups of benchmarks: LIBERO~\citep{liu2023libero}, SIMPLER~\citep{li2024simpler}, and a set of real-world robotic manipulation tasks, as shown in Figure~\ref{fig:Benchmark}. These benchmarks cover both simulation and real-robot settings, and include tasks with different horizons, scene complexity, and semantic requirements.

\paragraph{LIBERO.}
LIBERO~\citep{liu2023libero} is a language-conditioned manipulation benchmark built on Robosuite. We follow the standard four-suite protocol and evaluate on \textbf{Spatial}, \textbf{Object}, \textbf{Goal}, and \textbf{Long}, with 10 tasks in each suite. Spatial focuses on spatial relations under varying object configurations. Object changes the target object while keeping the task template relatively fixed. Goal evaluates goal-conditioned manipulation with different final states. Long contains multi-stage long-horizon tasks and is more sensitive to accumulated execution errors.

\paragraph{SIMPLER.}
SIMPLER~\citep{li2024simpler} is a simulation benchmark designed to better reflect real-robot manipulation performance. We evaluate on four Google-Robot-style tasks: \textbf{Pick Coke}, \textbf{Move Near}, \textbf{Open Drawer}, and \textbf{Close Drawer}. These tasks cover common short-horizon manipulation primitives, including grasping, local object repositioning, and articulated-object interaction. SIMPLER complements LIBERO by providing visually grounded tabletop tasks with a closer connection to real-robot evaluation.

\paragraph{Real-world tasks.}
We further conduct real-world experiments on an AgileX Piper robot. The real-world suite includes \textbf{Cube Grasping}, \textbf{Pick\&Place}, \textbf{Select-Pick\&Place}, and \textbf{Multi-Cube Stacking}. Cube Grasping evaluates basic target localization and grasp execution. Pick\&Place tests object transport and placement. Select-Pick\&Place introduces distractor objects and requires target selection before manipulation. Multi-Cube Stacking requires repeated grasping and placement, and is the most sequential task in this suite. These real-world tasks test whether aggressive visual-token pruning remains reliable under practical sensing and execution conditions.

\subsection{More results}
\label{sec:metrics}
\subsubsection{Controlled Comparison Under a Unified 32-Token Budget}
\label{app:libero_same_budget}

Table~\ref{tab:main_results} compares all applicable pruning methods on LIBERO under the same final budget of 32 visual tokens. SAPrune consistently achieves the strongest task performance across OpenVLA, OpenVLA-OFT, and $\pi_0$, indicating that its advantage comes from stage-aware pruning-layer placement and higher-quality token retention rather than from using more final tokens. SAPrune can be slightly slower than some baselines because it delays pruning until more reliable stages and reduces visual tokens progressively across selected layers, so earlier layers still process more tokens than early one-step pruning methods. Nevertheless, it preserves substantially higher success rates while still accelerating the full-token model, yielding the best overall efficiency-performance balance under the same token budget.

\begin{table}[H]
\centering
\small
\setlength{\tabcolsep}{4.0pt}
\renewcommand{\arraystretch}{1.12}
\caption{\textbf{Controlled comparison under a unified 32-token budget on LIBERO~\citep{liu2023libero}.}
All applicable pruning methods are evaluated with the same final retained-token budget of 32 visual tokens.
Red bold marks the best result among pruned methods within each backbone.}
\label{tab:main_results}

\resizebox{\textwidth}{!}{%
\begin{tabular}{lccccccccc}
\toprule
\multirow{2}{*}{Method} & \multirow{2}{*}{Tokens} & \multicolumn{5}{c}{LIBERO Success Rate (\%, $\uparrow$)} & \multirow{2}{*}{\makecell{Latency\\(ms, $\downarrow$)}} & \multirow{2}{*}{Speedup ($\uparrow$)} & \multirow{2}{*}{\makecell{FLOPs Ratio\\(\%, $\downarrow$)}} \\
\cmidrule(lr){3-7}
& & Spatial & Object & Goal & Long & Avg. & & & \\
\midrule

\rowcolor{sectiongray}
\multicolumn{10}{c}{\textbf{OpenVLA}~\citep{kim2025openvla}} \\
\midrule
OpenVLA           & 256 & 85.3 & 85.9 & 76.1 & 52.3 & 74.9 & 178.46 & 1.000$\times$ & 100.0 \\
+ FastV~\citep{chen2024fastv}            & 32  & 55.9 & 52.0 & 51.2 & 18.1 & 44.3 & 102.84 & 1.735$\times$ & 29.9 \\
+ DivPrune~\citep{alvar2025divprune}     & 32  & 50.1 & 49.8 & 47.8 & 14.9 & 40.7 & 103.76 & 1.720$\times$ & 33.0 \\
+ SP-VLA~\citep{li2025spvla}             & 32  & 68.0 & 69.5 & 63.4 & 30.0 & 57.7 & 101.32 & 1.761$\times$ & 32.1 \\
+ VLA-ADP~\citep{pei2025vlaadp}          & 32  & 72.2 & 68.4 & 62.9 & 34.0 & 59.4 & 100.66 & 1.773$\times$ & \best{28.4} \\
+ EfficientVLA(T-only)~\citep{yang2025efficientvla} & 32  & 70.6 & 69.0 & 60.4 & 31.6 & 57.9 & \best{99.58} & \best{1.792$\times$} & 32.0 \\
\rowcolor{oursbg}
+ \textbf{SAPrune}    & 32  & \best{84.9} & \best{85.1} & \best{75.3} & \best{50.8} & \best{74.0} & 110.96 & 1.608$\times$ & 39.5 \\

\midrule
\rowcolor{sectiongray}
\multicolumn{10}{c}{\textbf{OpenVLA-OFT}~\citep{kim2025openvlaoft}} \\
\midrule
OpenVLA-OFT           & 256 & 95.5 & 98.8 & 96.6 & 90.7 & 95.4 & 103.59 & 1.000$\times$ & 100.0 \\
+ FastV~\citep{chen2024fastv}            & 32  & 63.8 & 60.7 & 61.0 & 31.1 & 54.2 & 56.75 & 1.825$\times$ & 25.5 \\
+ DivPrune~\citep{alvar2025divprune}     & 32  & 60.9 & 62.2 & 62.9 & 29.3 & 53.8 & 55.88 & 1.854$\times$ & 27.9 \\
+ SP-VLA~\citep{li2025spvla}             & 32  & 72.0 & 65.3 & 71.8 & 43.7 & 63.2 & 54.23 & 1.910$\times$ & 26.3 \\
+ VLA-ADP~\citep{pei2025vlaadp}          & 32  & 80.3 & 73.5 & 79.4 & 51.7 & 71.2 & 54.71 & 1.893$\times$ & \best{24.6} \\
+ EfficientVLA(T-only)~\citep{yang2025efficientvla} & 32  & 76.2 & 74.9 & 77.1 & 54.2 & 70.6 & \best{53.68} & \best{1.930$\times$} & 25.7 \\
\rowcolor{oursbg}
+ \textbf{SAPrune}    & 32  & \best{95.2} & \best{98.1} & \best{95.9} & \best{88.9} & \best{94.5} & 60.30 & 1.718$\times$ & 37.8 \\

\midrule
\rowcolor{sectiongray}
\multicolumn{10}{c}{\bfseries\boldmath $\bm{\pi}_{0}$~\citep{black2024pi0}} \\
\midrule
\bfseries\boldmath $\bm{\pi}_{0}$           & 256 & 96.9 & 96.3 & 95.2 & 87.2 & 93.9 & 78.49 & 1.000$\times$ & 100.0 \\
+ FastV~\citep{chen2024fastv}            & 32  & 67.2 & 61.4 & 51.5 & 29.3 & 52.4 & 47.15 & 1.665$\times$ & \best{30.1} \\
+ DivPrune~\citep{alvar2025divprune}     & 32  & 60.0 & 65.5 & 64.8 & 32.0 & 55.6 & 47.53 & 1.651$\times$ & 35.0 \\
+ SP-VLA~\citep{li2025spvla}             & 32  & 78.2 & 69.7 & 60.6 & 47.8 & 64.1 & 46.66 & 1.682$\times$ & 32.2 \\
+ VLA-ADP~\citep{pei2025vlaadp}          & 32  & 78.0 & 77.8 & 80.0 & 56.3 & 73.0 & \best{46.18} & \best{1.700$\times$} & 31.4 \\
\rowcolor{oursbg}
+ \textbf{SAPrune}    & 32  & \best{96.2} & \best{95.8} & \best{94.4} & \best{86.0} & \best{93.1} & 51.22 & 1.532$\times$ & 38.6 \\
\bottomrule
\end{tabular}%
}
\end{table}

\subsubsection{Routing Overhead}
\label{app:routing}

Table~\ref{tab:routing} reports the wall-clock cost of Algorithm~\ref{alg:dual_path_pruning} at the three pruning stages.
The total routing overhead is 4.76\,ms on OpenVLA, with the functional-signature
construction and greedy coverage-selection component accounting for the largest
portion (3.09\,ms). The stage-wise cost decreases from 2.72\,ms to 1.35\,ms and
0.69\,ms as the visual-token set is progressively reduced. This indicates that
the additional computation introduced by the dual-path selection becomes
increasingly lightweight at later pruning stages.

\begin{table}[H]
\centering
\small
\setlength{\tabcolsep}{6pt}
\renewcommand{\arraystretch}{1.05}
\caption{\textbf{Wall-clock latency breakdown of Algorithm~2 on OpenVLA.} Latency is reported in milliseconds.}
\label{tab:routing}
\resizebox{\textwidth}{!}{%
\begin{tabular}{lrrrr}
\toprule
\textbf{Component} & \textbf{Shallow 256$\rightarrow$128} & \textbf{Middle 128$\rightarrow$64} & \textbf{Deep 64$\rightarrow$32} & \textbf{Total} \\
\midrule
Candidate ranking / Top-$M$ & 0.08 & 0.05 & 0.04 & 0.17 \\
Reliability computation ($P_l$, $A_l$, $\lambda_l$) & 0.49 & 0.32 & 0.17 & 0.98 \\
Functional signature + greedy selection & 1.90 & 0.82 & 0.37 & 3.09 \\
Token indexing / gather & 0.25 & 0.16 & 0.11 & 0.52 \\
\textbf{Algorithm~2 total} & \textbf{2.72} & \textbf{1.35} & \textbf{0.69} & \textbf{4.76} \\
\bottomrule
\end{tabular}
}
\end{table}

\Needspace{16\baselineskip}
\subsubsection{Attention Backend}
\label{app:backend}

Table~\ref{tab:backend} compares eager attention, PyTorch SDPA, and FlashAttention-2.
SAPrune achieves similar speedups across the three tested backends, showing that
its gains can be retained with standard attention optimizations.
\par\addvspace{\baselineskip}
\noindent\begin{minipage}{\linewidth}
\centering
\small
\setlength{\parskip}{0pt}
\setlength{\tabcolsep}{4pt}
\renewcommand{\arraystretch}{1.05}
\captionsetup{position=top,skip=7pt}
\captionof{table}{\textbf{Compatibility with standard attention backends.}}
\label{tab:backend}
\begin{tabular*}{\linewidth}{@{\extracolsep{\fill}}llrrr@{}}
\toprule
\textbf{Backbone} & \textbf{Attention backend} & \textbf{Full-token latency (ms)} & \textbf{SAPrune latency (ms)} & \textbf{Speedup} \\
\midrule
OpenVLA & Eager & 182.92 & 112.62 & 1.624$\times$ \\
OpenVLA & SDPA (default) & 178.46 & 110.96 & 1.608$\times$ \\
OpenVLA & FlashAttention-2 & 177.75 & 110.74 & 1.605$\times$ \\
OpenVLA-OFT & Eager & 105.87 & 61.20 & 1.730$\times$ \\
OpenVLA-OFT & SDPA (default) & 103.59 & 60.30 & 1.718$\times$ \\
OpenVLA-OFT & FlashAttention-2 & 103.18 & 60.18 & 1.715$\times$ \\
$\pi_0$ & Eager & 80.06 & 51.83 & 1.545$\times$ \\
$\pi_0$ & SDPA (default) & 78.49 & 51.22 & 1.532$\times$ \\
$\pi_0$ & FlashAttention-2 & 78.18 & 51.12 & 1.529$\times$ \\
\bottomrule
\end{tabular*}
\end{minipage}
\par\addvspace{\baselineskip}
\subsubsection{End-to-end latency on real robots.}
Table~\ref{tab:e2e_latency} provides a breakdown of the real-robot policy
pipeline latency on $\pi_0$. SAPrune reduces the VLA inference time from
78.16\,ms to 51.45\,ms, while the remaining pipeline components introduce
similar overhead across methods. This leads to an E2E latency reduction from
88.68\,ms to 61.97\,ms and increases the policy-pipeline frequency from 11.28\,Hz to
16.14\,Hz, showing that token pruning provides practical acceleration beyond
isolated model inference speedups.
\begin{table*}[!htbp]
\centering
\small
\setlength{\tabcolsep}{5pt}
\renewcommand{\arraystretch}{1.05}
\caption{\textbf{Real-robot policy-pipeline latency and E2E frequency on $\pi_0$.} E2E frequency is computed as $f_{\mathrm{E2E}}=1000/T_{\mathrm{E2E}}$. E2E latency also includes 0.61--0.62\,ms of orchestration and synchronization overhead.
Red bold marks the best result among pruned methods.}
\label{tab:e2e_latency}
\resizebox{\textwidth}{!}{%
\begin{tabular}{lcccccc}
\toprule
\textbf{Method} & \textbf{Preprocess (ms)} & \textbf{VLA inference (ms)} & \textbf{Action decoding (ms)} & \textbf{Communication (ms)} & \textbf{E2E latency (ms)} & \textbf{E2E frequency (Hz)} \\
\midrule
$\pi_0$~\citep{black2024pi0} & 6.08 & 78.16 & 2.46 & 1.36 & 88.68 & 11.28 \\
+ FastV~\citep{chen2024fastv} & 6.05 & 58.92 & 2.44 & 1.35 & 69.37 & 14.42 \\
+ SP-VLA~\citep{li2025spvla} & 6.04 & 58.40 & 2.47 & 1.37 & 68.89 & 14.52 \\
+ VLA-ADP~\citep{pei2025vlaadp} & 6.07 & 59.30 & 2.48 & 1.36 & 69.83 & 14.32 \\
\rowcolor{oursbg}
+ \textbf{SAPrune} & 6.10 & 51.45 & 2.46 & 1.35 & \best{61.97} & \best{16.14} \\
\bottomrule
\end{tabular}%
}
\end{table*}

\subsubsection{Calibration Transfer}
\label{app:calibration_transfer}

\begin{wraptable}{r}{0.50\columnwidth}
\centering
\small
\setlength{\tabcolsep}{3.5pt}
\renewcommand{\arraystretch}{0.95}
\caption{\textbf{Calibration transfer on OpenVLA.}
Calibration is performed on different data sources and evaluated on
LIBERO-Long without LIBERO-specific recalibration.}
\label{tab:calibration_transfer}
\resizebox{\linewidth}{!}{%
\begin{tabular}{lcc}
\toprule
\textbf{Calibration} & \textbf{Layers} & \textbf{Long SR (\%)} \\
\midrule
LIBERO (original)    & [2,16,26] & 50.8 \\
SIMPLER ($N=30$)     & [2,15,26] & 49.3 \\
Real-world ($N=60$)  & [2,17,26] & \textbf{51.1} \\
Real-world ($N=30$)  & [2,17,27] & 47.6 \\
\bottomrule
\end{tabular}%
}
\end{wraptable}

We further examine whether the calibration procedure transfers across
environments. Specifically, we calibrate OpenVLA using SIMPLER or real-world
rollouts and directly evaluate the resulting layer selections on
LIBERO-Long. No LIBERO calibration samples are used in these transfer
experiments, and all other settings are kept unchanged; in particular,
we perform neither retraining nor additional recalibration.

As shown in Table~\ref{tab:calibration_transfer}, calibration transferred
from SIMPLER or real-world data yields success rates of 47.6--51.1\%,
compared with 50.8\% when calibration is performed directly on LIBERO.
Notably, calibration with 60 real-world rollouts achieves 51.1\% success,
showing that the selected pruning layers can remain effective when the
calibration and evaluation environments differ. Here, $N$ denotes the
number of calibration rollouts, and all layer indices are 0-based.

When increasing the calibration set to 120 rollouts, LIBERO, SIMPLER,
real-world, and mixed-source data all select the same layers,
$[2,16,26]$. The mixed-source calibration also selects $[2,16,26]$ with
only 30 or 60 rollouts. These results indicate that, for a fixed model,
the selected pruning layers are relatively stable across the tested
calibration sources and can be reused across downstream tasks. A new
model still requires a separate calibration procedure, and transfer to
previously unseen settings should be validated empirically.
\subsubsection{Robustness on LIBERO-Plus}
\label{app:libero_plus}

We evaluate $\pi_{0.5}$ and SAPrune on LIBERO-Plus~\citep{fei2026liberoplus}
under seven distribution shifts. SAPrune is calibrated once on 300 trajectories
from standard LIBERO, with no LIBERO-Plus data used for calibration or
recalibration. Both methods are evaluated once on each of the same 10,030 task
variants under identical settings.

\begin{table}[H]
\centering
\small
\setlength{\tabcolsep}{4pt}
\renewcommand{\arraystretch}{1.05}
\caption{\textbf{Success rates (\%) under LIBERO-Plus distribution shifts.}
Overall is computed over all 10,030 task variants, accounting for different
category sizes. Tokens denotes visual tokens per image.}
\label{tab:libero_plus}
\resizebox{\linewidth}{!}{%
\begin{tabular}{lrrrrrrrrr}
\toprule
\textbf{Method} & \textbf{Tokens} & \textbf{Camera} & \textbf{Robot} & \textbf{Language} & \textbf{Light} & \textbf{Background} & \textbf{Noise} & \textbf{Layout} & \textbf{Overall} \\
\midrule
$\pi_{0.5}$~\citep{black2025pi05} & 256 & 75.7 & 77.1 & 85.9 & 96.4 & 94.9 & 90.1 & 85.2 & 85.7 \\
+ \textbf{SAPrune} & 32 & 74.5 & 76.1 & 84.8 & 95.6 & 93.8 & 88.6 & 83.9 & 84.5 \\
\bottomrule
\end{tabular}%
}
\end{table}

Table~\ref{tab:libero_plus} shows that SAPrune achieves 84.5\% overall success
with 32 tokens, compared with 85.7\% for the full model with 256 tokens.
The largest category gap is 1.5 percentage points under noise, indicating
only a small additional loss under the tested shifts.

\subsubsection{Performance under Different Final Prune Ratios}
\label{app:final_prune_ratio_analysis}
\begin{figure}[!htbp]
    \centering
    \begin{subfigure}[t]{0.24\textwidth}
        \centering
        \includegraphics[width=\textwidth]{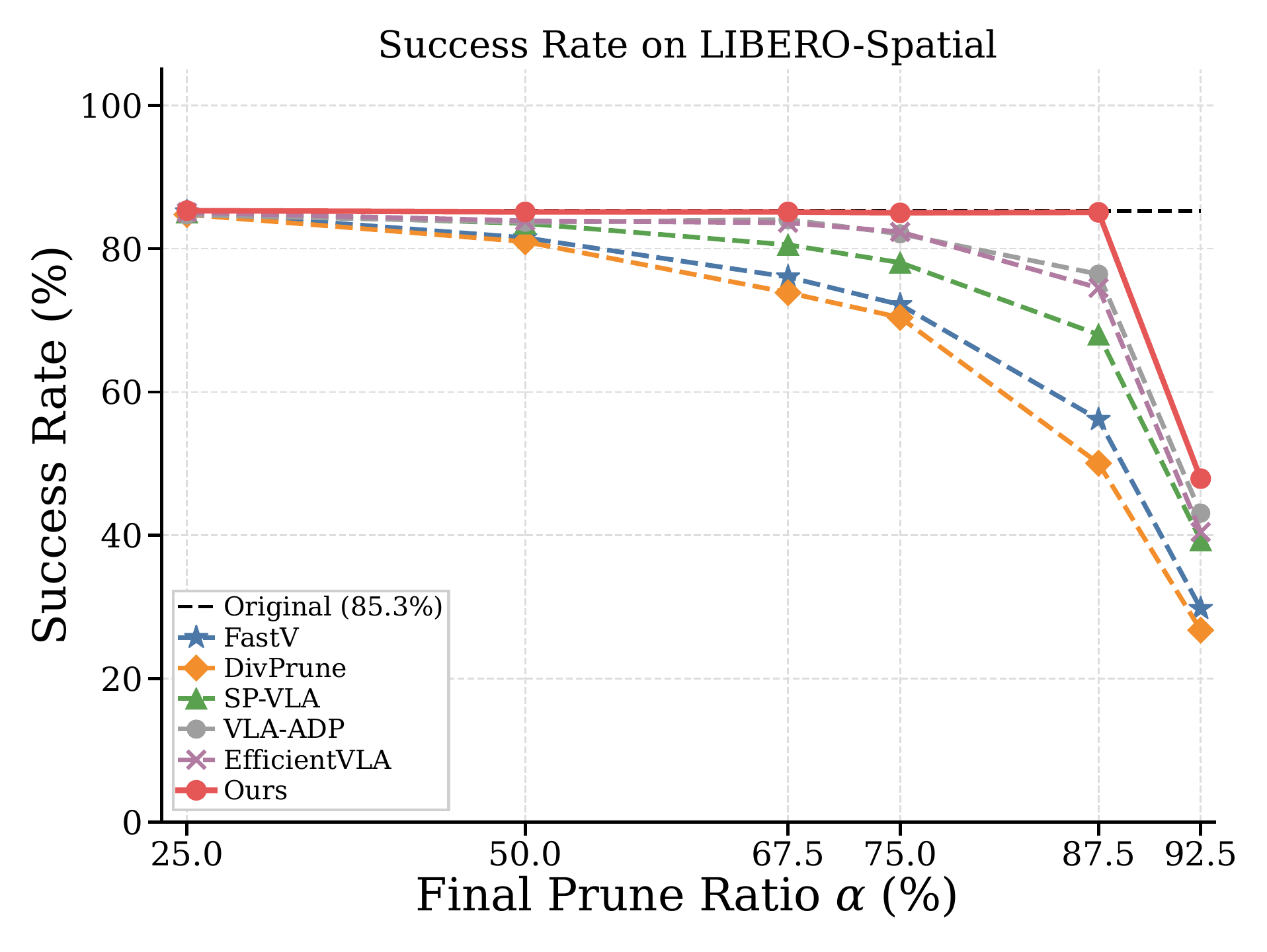}
        \caption{LIBERO-Spatial}
        \label{fig:OpenVLA_Spatial_curve}
    \end{subfigure}
    \hfill
    \begin{subfigure}[t]{0.24\textwidth}
        \centering
        \includegraphics[width=\textwidth]{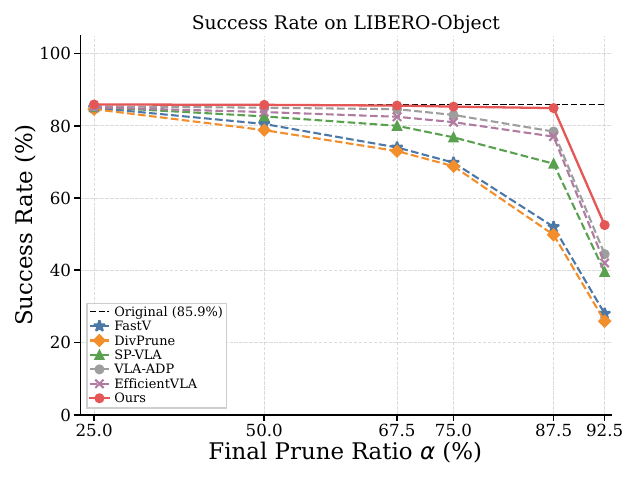}
        \caption{LIBERO-Object}
        \label{fig:OpenVLA_Object_curve}
    \end{subfigure}
    \hfill
    \begin{subfigure}[t]{0.24\textwidth}
        \centering
        \includegraphics[width=\textwidth]{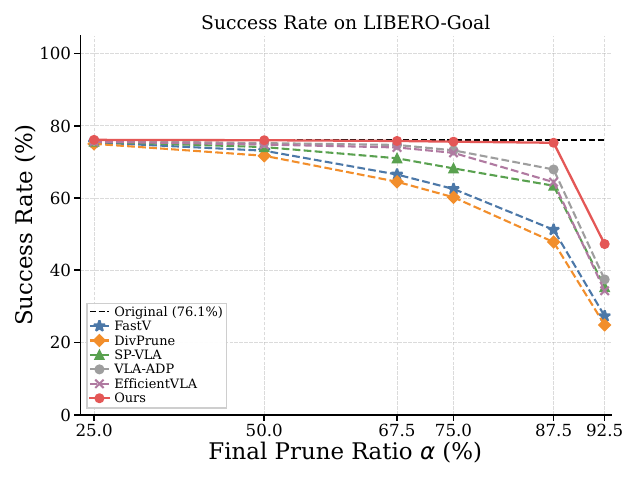}
        \caption{LIBERO-Goal}
        \label{fig:OpenVLA_Goal_curve}
    \end{subfigure}
    \hfill
    \begin{subfigure}[t]{0.24\textwidth}
        \centering
        \includegraphics[width=\textwidth]{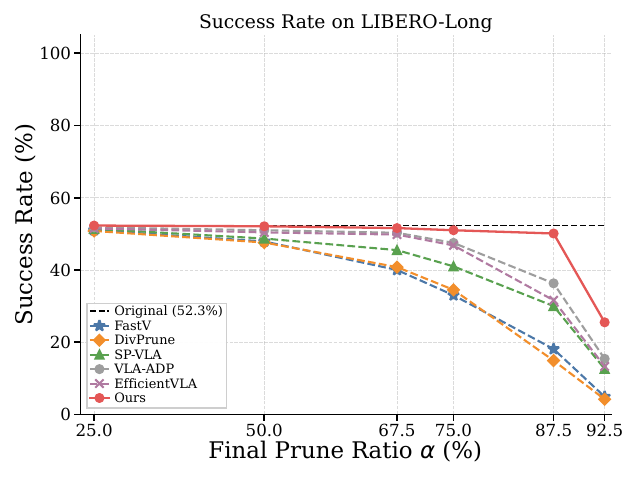}
        \caption{LIBERO-Long}
        \label{fig:OpenVLA_Long_curve}
    \end{subfigure}

    \caption{\textbf{Performance under different final prune ratios on OpenVLA.}
    Success rates on the four LIBERO suites are reported as the final prune ratio increases. Our method consistently preserves stronger performance, especially under aggressive pruning regimes.}
    \label{fig:openvla_prune_ratio_curves}
\end{figure}

\begin{figure}[!htbp]
    \centering
    \begin{subfigure}[t]{0.24\textwidth}
        \centering
        \includegraphics[width=\textwidth]{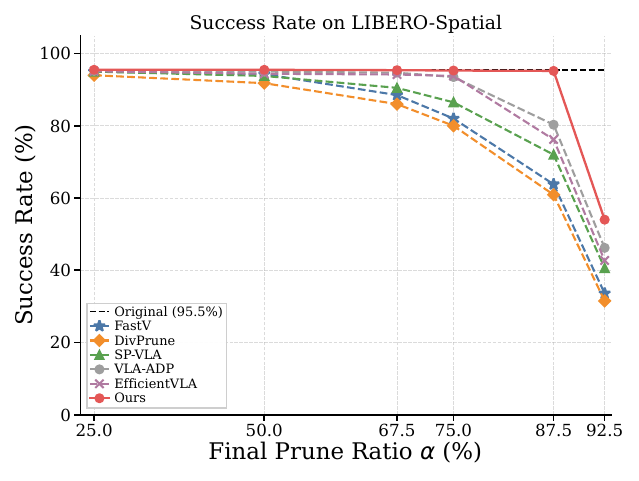}
        \caption{LIBERO-Spatial}
        \label{fig:OpenVLA-OFT_Spatial_curve}
    \end{subfigure}
    \hfill
    \begin{subfigure}[t]{0.24\textwidth}
        \centering
        \includegraphics[width=\textwidth]{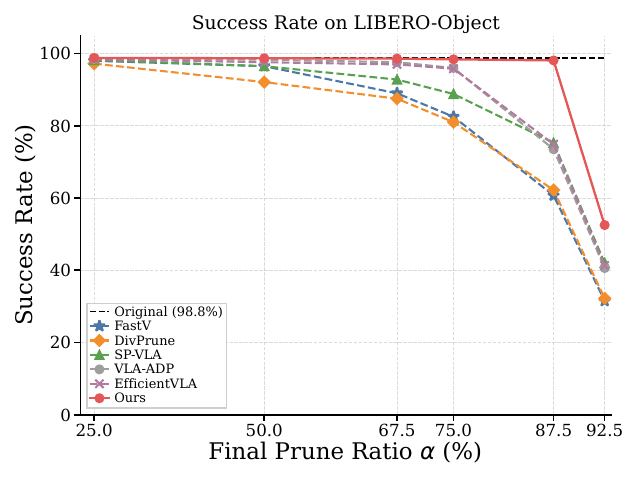}
        \caption{LIBERO-Object}
        \label{fig:OpenVLA-OFT_Object_curve}
    \end{subfigure}
    \hfill
    \begin{subfigure}[t]{0.24\textwidth}
        \centering
        \includegraphics[width=\textwidth]{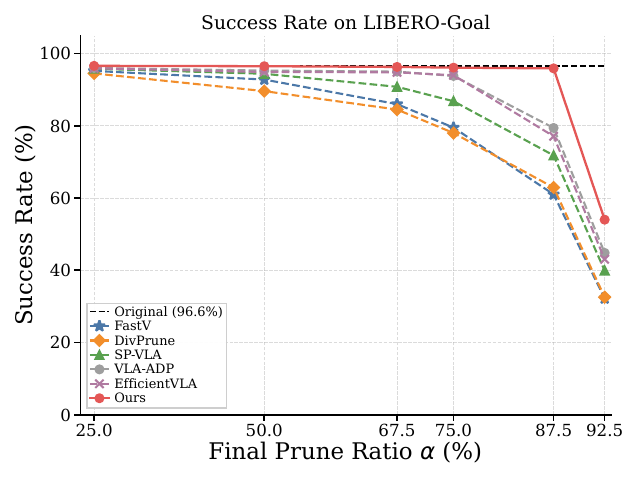}
        \caption{LIBERO-Goal}
        \label{fig:OpenVLA-OFT_Goal_curve}
    \end{subfigure}
    \hfill
    \begin{subfigure}[t]{0.24\textwidth}
        \centering
        \includegraphics[width=\textwidth]{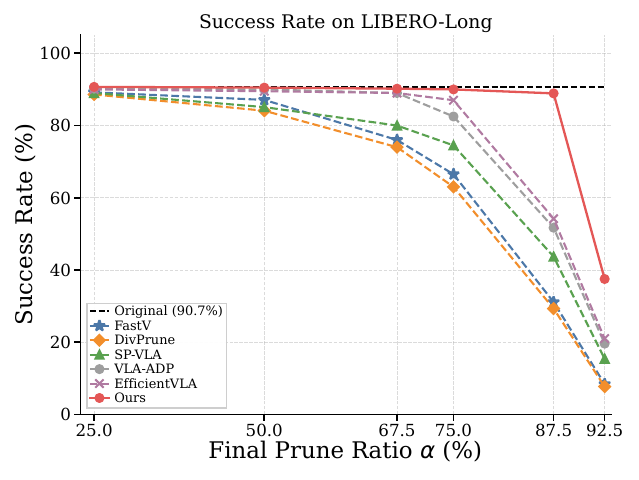}
        \caption{LIBERO-Long}
        \label{fig:OpenVLA-OFT_Long_curve}
    \end{subfigure}

    \caption{\textbf{Performance under different final prune ratios on OpenVLA-OFT.}
    Success rates on the four LIBERO suites are reported as the final prune ratio increases. Our method remains the most robust among the compared pruning methods across a wide range of pruning levels.}
    \label{fig:openvla_oft_prune_ratio_curves}
\end{figure}

\begin{figure}[!htbp]
    \centering
    \begin{subfigure}[t]{0.24\textwidth}
        \centering
        \includegraphics[width=\textwidth]{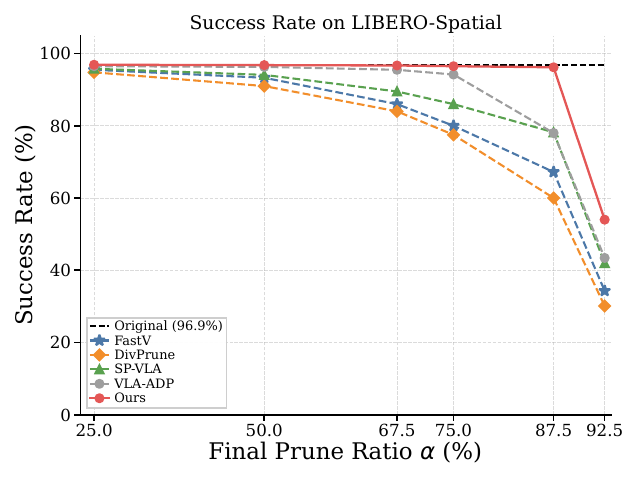}
        \caption{LIBERO-Spatial}
        \label{fig:pi0_Spatial_curve}
    \end{subfigure}
    \hfill
    \begin{subfigure}[t]{0.24\textwidth}
        \centering
        \includegraphics[width=\textwidth]{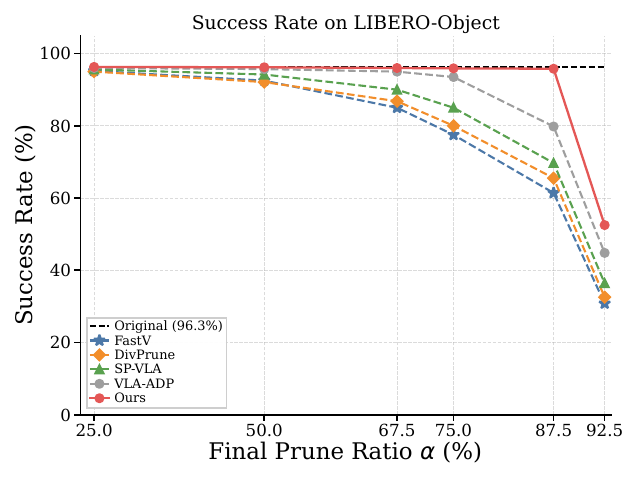}
        \caption{LIBERO-Object}
        \label{fig:pi0_Object_curve}
    \end{subfigure}
    \hfill
    \begin{subfigure}[t]{0.24\textwidth}
        \centering
        \includegraphics[width=\textwidth]{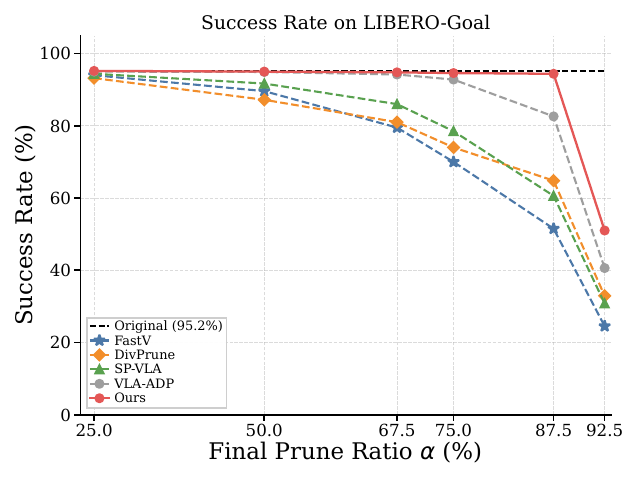}
        \caption{LIBERO-Goal}
        \label{fig:pi0_Goal_curve}
    \end{subfigure}
    \hfill
    \begin{subfigure}[t]{0.24\textwidth}
        \centering
        \includegraphics[width=\textwidth]{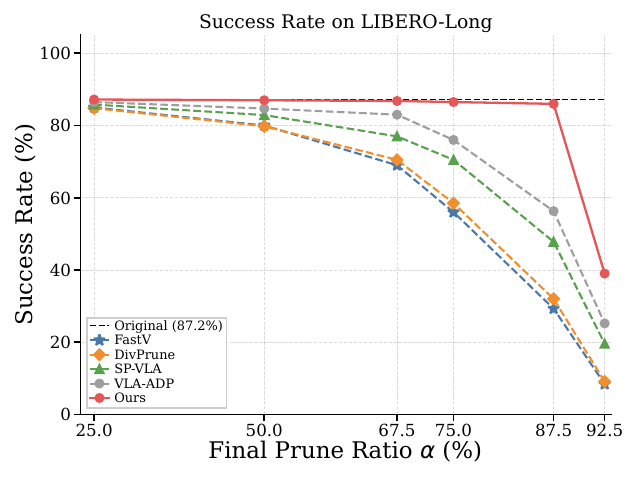}
        \caption{LIBERO-Long}
        \label{fig:pi0_Long_curve}
    \end{subfigure}

    \caption{\textbf{Performance under different final prune ratios on $\pi_0$.}
    Success rates on the four LIBERO suites are reported as the final prune ratio increases. The results show that our method better maintains policy performance as pruning becomes more aggressive.}
    \label{fig:pi0_prune_ratio_curves}
\end{figure}

Figures~\ref{fig:openvla_prune_ratio_curves}, \ref{fig:openvla_oft_prune_ratio_curves}, and \ref{fig:pi0_prune_ratio_curves} report the success rates on the four LIBERO suites as the final prune ratio increases on OpenVLA, OpenVLA-OFT, and $\pi_0$, respectively. Across all three backbones, a consistent trend can be observed: as pruning becomes more aggressive, the performance of all methods gradually degrades, with the most pronounced drops appearing in the high-pruning regime. Compared with the competing baselines, \textbf{SAPrune} maintains stronger performance over a wide range of prune ratios and shows better robustness especially under aggressive token reduction.

\subsection{More Ablation}

Table~\ref{tab:retention_path_ablation} ablates the two token-retention paths under the same pruning layers and 32-token budget. Removing either path consistently degrades performance, showing that both components are necessary for aggressive visual-token pruning. The attention-core path contributes the larger gain, indicating that highly action-attended tokens are the most reliable evidence for control. The functional-diversity path still provides clear additional benefit by preserving complementary context that attention-only selection may miss, such as surrounding objects, spatial anchors, and instruction-conditioned cues. 

\begin{table}[H]
\centering
\tiny
\setlength{\tabcolsep}{5pt}
\renewcommand{\arraystretch}{1}
\caption{\textbf{Ablation study of token-retention paths on LIBERO~\citep{liu2023libero}.}
All variants use the same stage-aware pruning layers and the same 32-token budget; only the retention rule is changed.
Red bold marks the best success rate among pruning variants.}
\label{tab:retention_path_ablation}
\resizebox{\columnwidth}{!}{%
\begin{tabular}{lccccccc}
\toprule
\multirow{2}{*}{Variant}
& \multicolumn{5}{c}{LIBERO Success Rate (\%, $\uparrow$)}
& \multirow{2}{*}{\makecell{Latency\\(ms, $\downarrow$)}}
& \multirow{2}{*}{\makecell{FLOPs Ratio\\(\%, $\downarrow$)}} \\
\cmidrule(lr){2-6}
& Spatial & Object & Goal & Long & Avg. & & \\
\midrule
Full SAPrune
& \best{84.9} & \best{85.1} & \best{75.3} & \best{50.8} & \best{74.0}
& 110.96 & 39.5 \\

w/o Attention core
& 75.9 & 75.1 & 65.7 & 39.0 & 63.9
& 111.47 & 40.2 \\

w/o Functional diversity
& 78.1 & 77.6 & 68.9 & 43.2 & 67.0
& 107.93 & 39.1 \\
\bottomrule
\end{tabular}%
}

\vspace{2pt}

\end{table}

\subsubsection{More analysis}
\label{app:gamma_sensitivity}

Figure~\ref{fig:poolab} shows the effect of the candidate-pool expansion coefficient $\gamma$ on the average LIBERO success rate. When $\gamma$ is too small, the candidate pool is overly restrictive and may exclude potentially useful visual tokens, leading to clear performance degradation. As $\gamma$ increases, the performance improves steadily, since a larger candidate pool provides more room for complementary evidence selection. The curve reaches its peak around $\gamma \approx 1.75$ and then remains relatively stable, suggesting that once the candidate pool is sufficiently expressive, further expansion brings only marginal benefit. Based on this observation, we use $\gamma=1.75$ in our experiments.

\subsubsection{Sensitivity and allocation analysis.}
\begin{figure}[!htbp]
    \centering

    \begin{subfigure}[t]{0.39\textwidth}
        \centering
        \includegraphics[width=\textwidth]{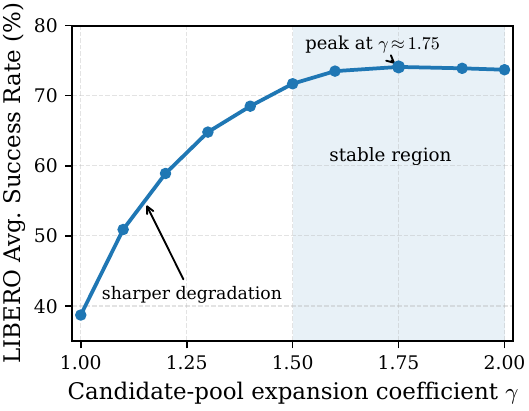}
        \caption{Sensitivity to candidate-pool size}
        \label{fig:poolab}
    \end{subfigure}
    \hfill
    \begin{subfigure}[t]{0.601\textwidth}
        \centering
        \includegraphics[width=\textwidth]{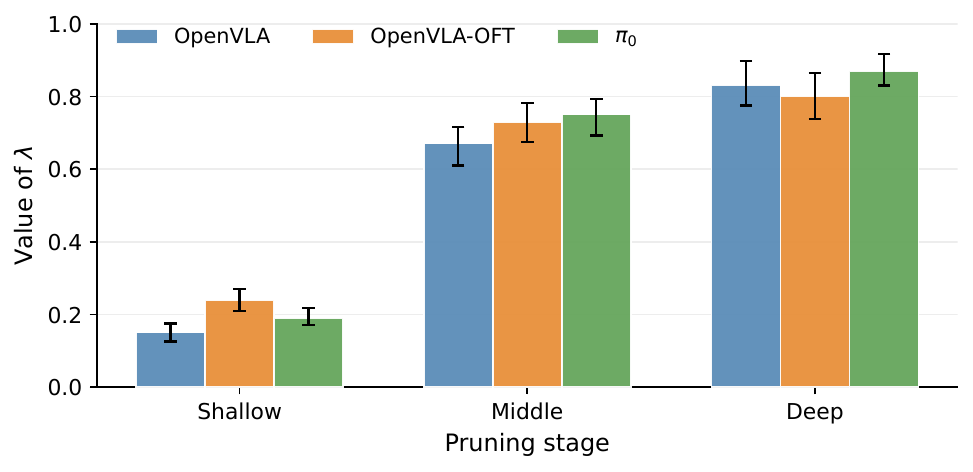}
        \caption{Layer-wise allocation coefficient}
        \label{fig:lambda_allocation}
    \end{subfigure}

    \caption{
    Sensitivity and allocation analysis of SAPrune.
    (a) The average LIBERO success rate increases with the candidate-pool expansion coefficient $\gamma$ and becomes stable when $\gamma$ is moderately large.
    (b) The layer-wise allocation coefficient $\lambda$ remains small in shallow layers and increases in middle and deep layers.
    }
    \label{fig:sensitivity_allocation_analysis}
\end{figure}

Figure~\ref{fig:sensitivity_allocation_analysis} shows two overall trends. First, the average LIBERO success rate increases as the candidate-pool expansion coefficient $\gamma$ grows, and then gradually enters a stable region when $\gamma$ becomes moderately large. Second, the allocation coefficient $\lambda$ shows a clear depth-wise increasing pattern across different VLA backbones: it remains small in shallow layers, becomes larger in middle layers, and reaches the highest values in deep layers.

\section{Limitation}
\label{sec:limitation}

SAPrune is built on the empirically supported premise that action-to-visual attention can serve as a useful proxy for action-relevant visual evidence. 
Our experiments suggest that this premise holds well for sufficiently trained VLA backbones whose action-related states are grounded in task-relevant visual regions. 
Under such conditions, SAPrune is better aligned with VLA inference than generic visual-token pruning methods because it delays pruning until action-conditioned attention becomes stable and preserves both dominant and complementary visual evidence. 
However, this premise may become weaker when the backbone attention is poorly grounded, remains diffuse, or is dominated by spurious background regions. 
In such cases, the attention signal may not accurately reflect the causal contribution of visual tokens, and SAPrune may still introduce performance degradation, especially in scenarios involving out-of-distribution objects, fine-grained contact regions, heavy occlusion, or long-horizon manipulation.

SAPrune also requires visual and action-related representations to interact
within the Transformer being pruned. This holds for the implementations
evaluated here. In separated vision-language model (VLM) and diffusion
Transformer (DiT) systems, action-conditioned queries may be available only
in the downstream action module. Direct use in the VLM then requires an
adaptation across modules, which we have not evaluated. Pruning the VLM
conditioning sequence would not remove the remaining DiT denoising computation,
limiting the possible end-to-end speedup.

\section{Impact Statement}
This work aims to improve the efficiency of vision-language-action models by reducing visual-token computation during inference. More efficient VLA inference can lower latency and deployment cost, potentially making robotic manipulation systems more practical on resource-constrained hardware. At the same time, efficiency improvements may also make robotic automation easier to deploy at larger scale, which raises broader concerns about safety, accountability, and labor displacement. Since token pruning may remove visual evidence needed for reliable control in rare or out-of-distribution situations, SAPrune should be used with appropriate validation, monitoring, and task-specific safety constraints before real-world deployment.

\section{ More Visualization}
\begin{figure}[!htbp]
    \centering
    \includegraphics[width=0.9\linewidth]{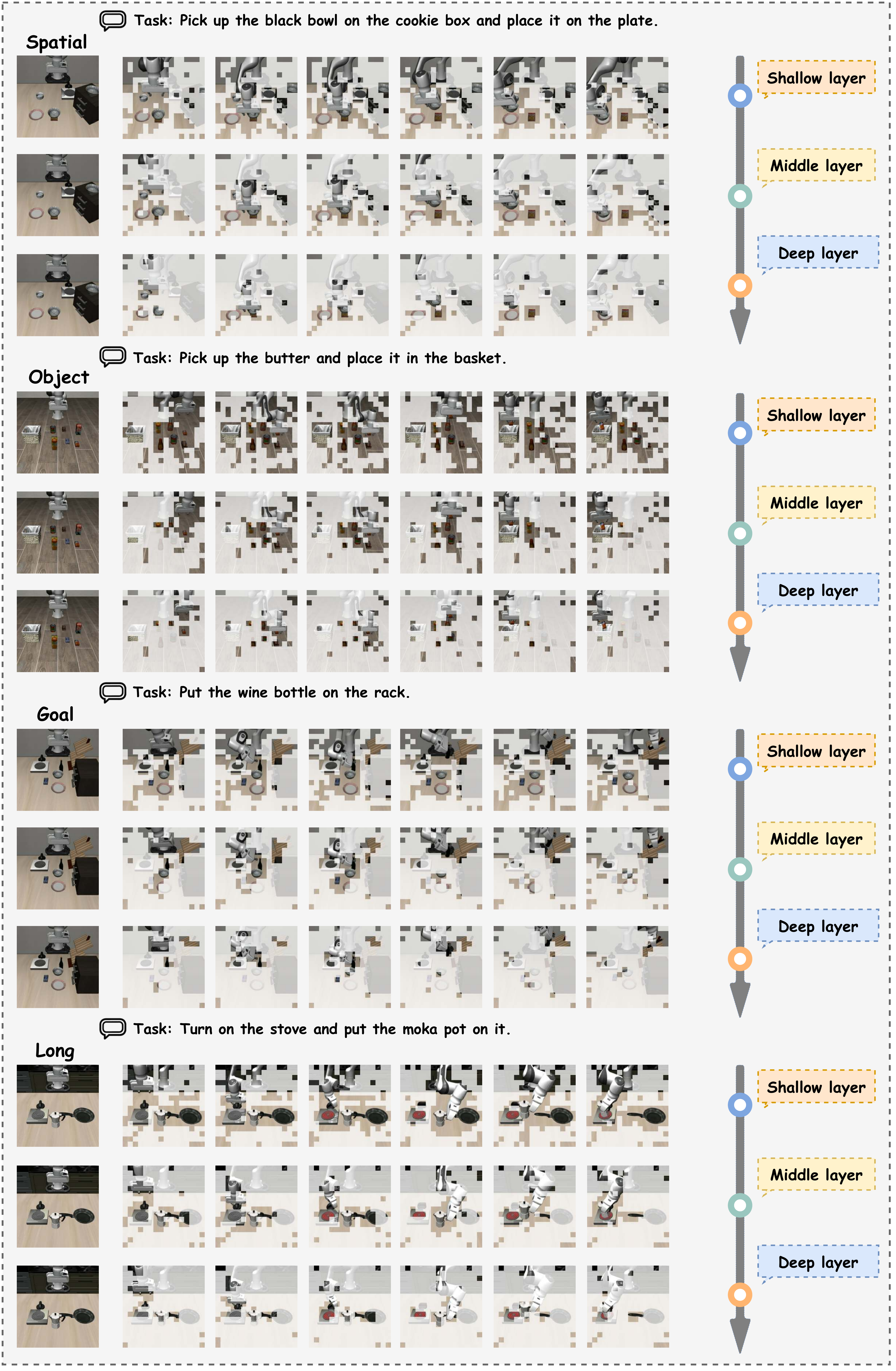}
    \caption{Visualization of SAPrune token pruning results across LIBERO tasks .}
    \label{fig:attention_dynamics_prune}
\end{figure}

\begin{figure}[!htbp]
    \centering
    \includegraphics[width=\linewidth]{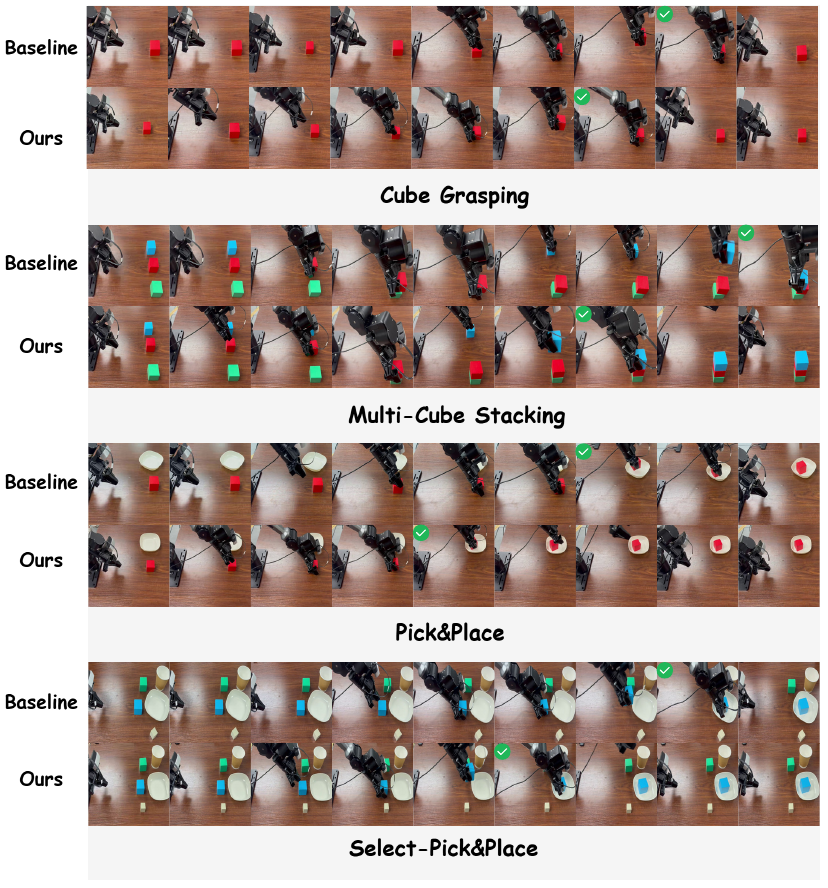}
    \caption{{\textbf{Real-world rollout comparison between the full $\pi_0$ baseline and SAPrune.}
We compare baseline and SAPrune rollouts on four real-robot manipulation tasks, including cube grasping, pick-and-place, multi-cube stacking, and select-pick-and-place. Green check marks indicate successful task completion.}}
    \label{fig:real_agx}
\end{figure}
\noindent\textbf{Real-World Robot Evaluation.}
Figure~\ref{fig:real_agx} shows qualitative real-world rollout comparisons between the full $\pi_0$ baseline and SAPrune on four manipulation tasks. 
Across cube grasping, pick-and-place, multi-cube stacking, and select-pick-and-place, SAPrune maintains similar execution behavior to the baseline under aggressive visual token pruning.

\noindent\textbf{Qualitative Analysis.}
Figure~\ref{fig:attention_dynamics_prune} visualizes SAPrune token pruning results across representative tasks from the LIBERO Spatial, Object, Goal, and Long suites.
The retained visual tokens remain concentrated around task-relevant objects and interaction regions across shallow, middle, and deep pruning stages, indicating that SAPrune preserves action-critical visual evidence under aggressive pruning.

\end{document}